\documentclass{article}

\usepackage{arxiv}
\usepackage[utf8]{inputenc} 
\usepackage[T1]{fontenc}    
\usepackage{hyperref}       
\usepackage{url}            
\usepackage{booktabs}       
\usepackage{amsfonts}       
\usepackage{nicefrac}       
\usepackage{microtype}      
\usepackage{graphicx}
\usepackage{natbib}
\usepackage{doi}

\usepackage{siunitx}		
\usepackage{multicol} 		
\usepackage{multirow} 		
\usepackage{xcolor}			
\definecolor{tab10blue}{RGB}{31, 119, 180}
\definecolor{tab10orange}{RGB}{255, 127, 14}
\definecolor{tab10green}{RGB}{44, 160, 44}
\definecolor{tab10red}{RGB}{214, 39, 40}
\usepackage{amsmath}       	
\usepackage{adjustbox}      
\usepackage{cleveref}		
\usepackage[super]{nth}		
\usepackage{nomencl}		
\makenomenclature

\title{Deep particulate matter forecasting model using correntropy-induced loss}

\author{ \href{https://orcid.org/0000-0001-7094-5518}{\includegraphics[scale=0.06]{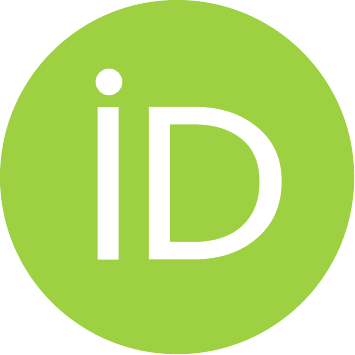}\hspace{1mm}Jongsu Kim} \\
	School of Mathematics and Computing\\
	Yonsei University\\
	Seoul, 03722 \\
	\texttt{skyobserver@yonsei.ac.kr} \\
	\And
	\href{https://orcid.org/0000-0002-9813-8117}{\includegraphics[scale=0.06]{orcid.pdf}\hspace{1mm}Changhoon Lee} \\
	Department of Mechanical Engineering\\
	Yonsei University\\
	Seoul, 03722 \\
	\texttt{clee@yonsei.ac.kr} \\
}

\renewcommand{\shorttitle}{Deep particulate matter forecasting model using correntropy-induced loss}

\hypersetup{
pdftitle={Deep particulate matter forecasting model using correntropy-induced loss},
pdfsubject={cs.CE, physics.ao-ph},
pdfauthor={Jongsu Kim, Changhoon Lee},
pdfkeywords={Air Pollution, Deep Learning,
Particulate Matter, Time Series Forecasting,
Heavy-Tailed Distribution, Correntropy},
}

\begin{document}
\maketitle

\begin{abstract}
Forecasting the particulate matter (PM) concentration
in South Korea has become urgently necessary
owing to its strong negative impact on human life.
In most statistical or machine learning methods,
independent and identically distributed data,
for example, a Gaussian distribution, are assumed;
however, time series such as air pollution and weather data
do not meet this assumption.
In this study,
the maximum correntropy criterion for regression (MCCR) loss
is used in an analysis of the statistical characteristics
of air pollution and weather data.
Rigorous seasonality adjustment
of the air pollution and weather data was performed
because of their complex seasonality patterns
and the heavy-tailed distribution of data
even after deseasonalization.
The MCCR loss was applied to multiple models
including conventional statistical models and
state-of-the-art machine learning models.
The results show that the MCCR loss is more appropriate
than the conventional mean squared error loss
for forecasting extreme values.
\end{abstract}


\section{Introduction}
\label{sec:introduction}
In recent years, air pollution has become a serious problem in South Korea.
Among air pollutants, particulate matter (PM) is the greatest concern.
PM\textsubscript{10} ($\leq 10$ \si{\micro\metre} in size) and
PM\textsubscript{2.5} ($\leq 2.5$ \si{\micro\metre} in size)
are widely known to cause health problems such as cardiovascular and/or
respiratory diseases \citet{EPA2016,Kim2018,Pope2006}. At high concentrations,
PM\textsubscript{2.5} not only causes health problems,
but also reduces agricultural productivity and increases socioeconomic costs \citet{OECD2006}.
Therefore, it is necessary to develop a timely and accurate
air pollution forecasting model that can operate in dynamic environments.

In South Korea, the yearly average concentrations of
PM\textsubscript{10} and PM\textsubscript{2.5} are
41 and 23 \si{\micro\gram\per\cubic\metre} \citet{KoreaEnvironmentCorporation2019},
which exceed the WHO air quality guidelines of
20 and 10 \si{\micro\gram\per\cubic\metre} \citet{WorldHealthOrganization2006}, respectively.
In addition, the yearly average concentration of PM\textsubscript{2.5}
is the highest among 37 OECD countries and
is almost twice the average value for OECD countries,
13.93 \si{\micro\gram\per\cubic\metre} \citet{OECD2021}.

Despite the seriousness of the situation,
there are many obstacles to the forecasting of PM\textsubscript{10} and PM\textsubscript{2.5}.
First, the concentrations of PM\textsubscript{10} and PM\textsubscript{2.5}
are affected by various factors,
such as anthropogenic emissions and meteorological parameters \citet{Makar2015}.
It is impossible to parameterize all of these factors,
especially in real time; moreover,
even though it is possible for some factors,
there is still considerable uncertainty in the models and observation data \citet{Bouarar2019}.
Second, spatiotemporal air pollution data for South Korea are limited.
South Korea is located in Northeast Asia, which has undergone rapid industrialization,
and it remains one of the most severely polluted regions worldwide
despite numerous efforts to address this issue \citet{OECD2006}.
There is an urgent demand for the development of a method of
forecasting the spatiotemporal concentrations of PM\textsubscript{10} and PM\textsubscript{2.5}.
Moreover, recent studies showed that the amount of PM\textsubscript{10} and PM\textsubscript{2.5}
contributed by domestic sources is also significant \citet{Lee2019,Jordan2020},
and the quantitative effect of air pollution from China is still controversial.

Many researchers have proposed various forecasting models to predict PM concentrations accurately.
Deterministic and statistical methods are generally used
for PM\textsubscript{10} and PM\textsubscript{2.5} prediction \citet{Singh2012, Zhang2012}.
In a deterministic model, physical or chemical equations are solved
to model the creation and transport of pollutants.
The most common deterministic models are CMAQ \citet{Byun1999}, GEOS-Chem \citet{Bey2001}, and WRF/Chem \citet{Grell2005}.
This approach has achieved great success because the models yield consistent and accurate results.
However, they require extensive observation data,
an accurate understanding of the physical interactions between inputs,
the correct representation of meteorological processes,
and appropriate assumptions regarding default parameters,
in particular anthropogenic emissions.
They are computationally expensive,
and thus a supercomputing system is needed for daily forecasting \citet{Baklanov2008,Zhou2017}.

By contrast, statistical models have been developed
to overcome these disadvantages of deterministic models.
For example, classification and regression trees (CART) using decision tree \citet{Shang2019}
and fuzzy logic for parameter classification \citet{Guo2007} are used to forecast air pollution statistically.
Time series forecasting models such as autoregressive integrated moving average (ARIMA) models
are also applied in air pollution prediction \citet{Wang2009}.
These models are supported by a well-established statistical and theoretical background
and are sometimes more accurate for site-specific prediction.
However, a major limitation of statistical models is
that they cannot represent physical or chemical processes directly;
thus, their results are sometimes unreliable.

The great success of deep learning models has attracted much attention recently.
It has been proved that deep learning models can be applied to
not only image processing and natural language processing,
but also time series forecasting \citet{Lai2018,Shih2019,Li2019}.
On the basis of advances in deep learning methodology,
several attempts have been made to apply deep learning models
to air pollution prediction modeling \citet{Choi2018,Cho2019,Franceschi2018,Bai2019}.
Although these studies showed promising results,
they focused on short-term prediction; consequently, long-term ($\geq 12$ h) results are uncertain.
In addition, they used various architectures based on a multilayer per-ceptron (MLP)
or long short-term memory (LSTM), which cannot easily capture complex patterns in data.
The increasing complexity of the models has been the subject of much research in recent years.
Unfortunately, the importance of the data distribution is sometimes ignored.
Deep learning models assume independent and identically distributed (i.i.d.) random variables or a Gaussian distribution;
however, real-world time series data are frequently dependent data
and have a non-Gaussian distribution.

To reflect the data distribution and capture complex nonlinear patterns,
a model with maximum correntropy criterion induced losses \citet{Liu2007,Feng2015}
is proposed in this study.
Environmental data, including air pollution and weather data,
are known to have long-term seasonal variation and heavy-tailed distributions \citet{Cichowicz2017}.
We take into account the non-Gaussian characteristics of the data
and develop a methodology that reflects these characteristics
and offers improved prediction accuracy compared to other forecasting methods.
Various machine learning and statistical models are tested.
In addition, the model performance for real-world data is analyzed,
and their limitations are also investigated.
The data preprocessing and data collection are described in detail in \autoref{sec:materials-analysis}.
The methods of the machine learning models are explained in detail in \autoref{sec:method},
and \autoref{sec:results} presents the model results and analysis. Conclusions are presented in \autoref{sec:conclusion}.

\section{Materials Analysis}
\label{sec:materials-analysis}
\subsection{Data Collection and Study Area}
\label{subsec:data-collection}

Two types of observation data,
air quality data and weather data,
are selected as explanatory variables,
and PM\textsubscript{10} and PM\textsubscript{2.5} are selected as the targets of our model.
Air quality data were collected in Jongno District in the central region of Seoul, the capital of South Korea.
Jongno District is chosen as the target site because it is located in the center of Seoul.
Weather data were measured at Seoul weather station,
where the Jongno air quality observatory is located.
Air quality data are publicly available at Airkorea (\url{https://www.airkorea.or.kr/eng}),
and weather data are available from the Korea Meteorological Administration (\url{https://data.kma.go.kr}).
Air quality data for Seoul from 2008 to
2014 are available only from the Seoul metropolitan government by request.
The annual mean temperature and precipitation of Seoul from 1980 to 2010 were
12.5\si{\celsius} and 1450\si{\milli\metre}.
However, the monthly average temperature varies from -2.4\si{\celsius} to 25.7\si{\celsius},
and 61\si{\%} of the annual precipitation falls in three months (June, July, and August).
South Korea is in the mid-latitudes of the Northern Hemisphere and
is affected by seasonal prevailing surface winds in the southwesterly
and northwesterly directions in summer and winter, respectively.
The seasonal variation in air quality and weather data has been extensively analyzed \citet{Cichowicz2017}.

The total data set covers January 2008 to October 2020 with hourly temporal resolution.
The training set data cover January 2008 to December 2018,
and the test set covers January 2019 to October 2020.
All missing data were interpolated
using the k-nearest neighbor algorithm implemented in scikit-learn \citet{Troyanskaya2001,Pedregosa2011}.
The wind direction in the raw data is given using 16 cardinal directions,
which were converted to degrees and encoded using a trigonometric function
because of the cyclical nature of the data.
The input data are standardized without the standard de-viation after seasonal adjustment,
as described \autoref{subsec:seasonal-adjustment}.

\begin{table}

	\caption{Feature information}

	\centering
	\begin{tabular}{lcc}
		\toprule
		Input parameters            &  Unit         & Category \\
		\midrule
		SO\textsubscript{2}         &  \si{ppm}     & \multirow{6}{*}{Air Quality}    \\
		CO                          &  \si{ppm}     & \\
		NO\textsubscript{2}         &  \si{ppm}     & \\
		PM\textsubscript{10}        &  \si{\micro\gram\per\cubic\metre} & \\
		PM\textsubscript{2.5}       &  \si{\micro\gram\per\cubic\metre} & \\
		\midrule
		Temperature                 &  \si{\celsius}                    & \multirow{7}{*}{Weather} \\
		Wind speed                  &  \si{\metre\per\second}           & \\
		Wind direction (sin)   		&                                   & \\
		Wind direction (cos)     	&                                   & \\
		Ground level pressure       &  \si{\hecto\pascal}               & \\
		Relative humidity           &  \%                               & \\
		Precipitation               &  \si{\milli\metre}                & \\
		\midrule
		Output parameters           &  Unit                             & Category \\
		\midrule
		PM\textsubscript{10}        &  \si{\micro\gram\per\cubic\metre} & \multirow{2}{*}{Air Quality} \\
		PM\textsubscript{2.5}       &  \si{\micro\gram\per\cubic\metre} &   \\
		\bottomrule
	\end{tabular}
	\label{tab:feature}
\end{table}

\subsection{Seasonal Adjustment of Data}
\label{subsec:seasonal-adjustment}

Most time series are composed of signal and noise.
Time series forecasting models are intended to capture the signal obscured by noise.
A time series model learns signals from the past under the assumption
that the variables are i.i.d., which introduces mixing and stationarity.
Mixing represents asymptotic independence;
that is, time series dependence between lags goes to zero as the lag increases.
Stationarity indicates that random variables are identically distributed.
The statistical properties of stationary time series
such as mean, variance, and autocorrelation are invariant over time.
However, time series are serially dependent by nature;
thus, the assumptions of stationarity and mixing are, strictly speaking, impossible.
Therefore, the effective sample size,
which represents the same statistical quantity even for a smaller sample size,
ust be determined by investigating the autocorrelation.

If a time series has a trend or seasonality, it is not considered a stationary time series.
As noted in \autoref{sec:introduction},
environmental variables, air pollution, and weather variables exhibit seasonal variation.
In this section, we describe how the seasonality was removed from the observation data
and examine the stationarity.

\begin{figure}
	\centering
	\includegraphics[width=\textwidth]{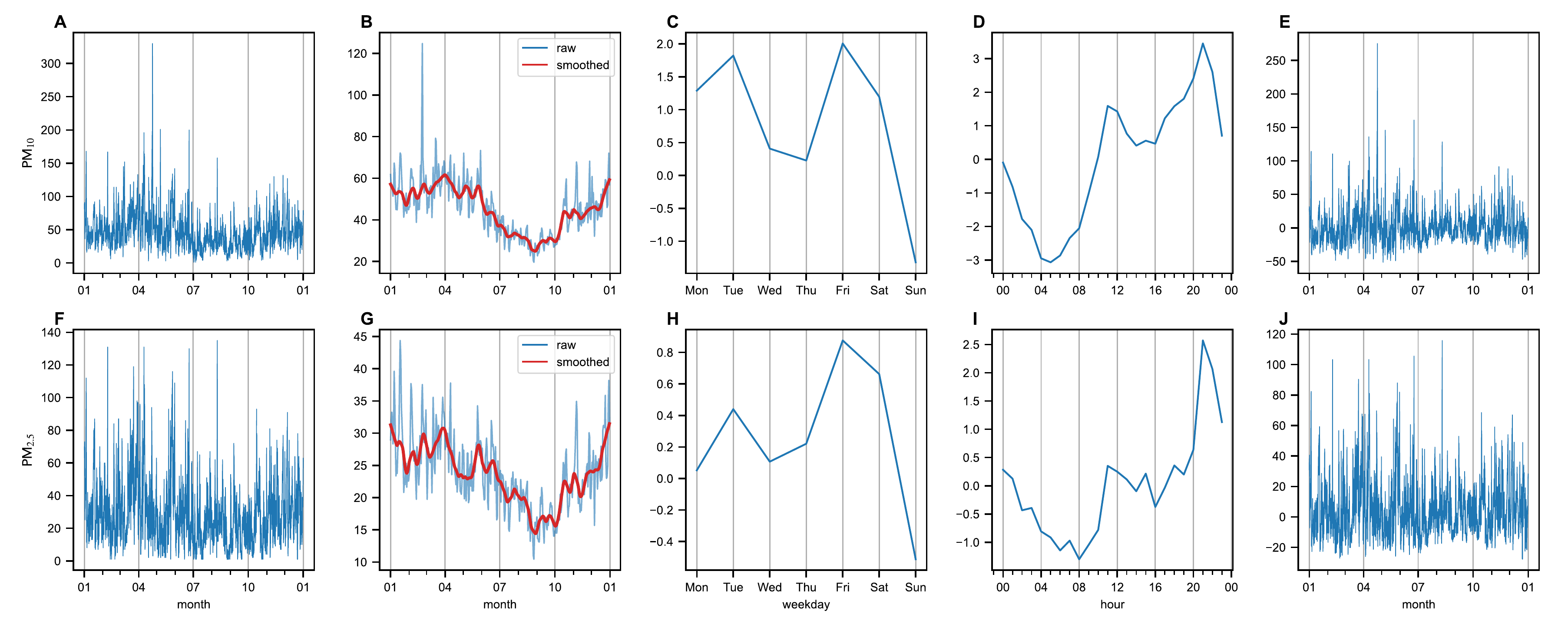}
	\caption{Seasonality and residuals of PM\textsubscript{10} and PM\textsubscript{2.5}.
		\textbf{A, F}: raw data, \textbf{B, G}: yearly seasonality
		(\textcolor{tab10red}{red} and \textcolor{tab10blue}{blue}
		represent seasonality with and without smoothing, respectively),
		\textbf{C, H}: weekly seasonality, \textbf{D, I}: daily seasonality, \textbf{E, J}: residuals}
	\label{fig:seasonality}
\end{figure}

Seasonal adjustment, or deseasonalization, is described by the following equation.
\begin{equation}
	\label{eq:seasonality}
	x(t) \approx s_{y,\textrm{smoothed}} + s_w + s_h + \textrm{res}_h
\end{equation}
where original series $x(t)$
is approximated by the sum of the smoothed yearly seasonality $s_{y,smoothed}$,
weekly variation $s_w$, daily variation $s_h$, and residual $\textrm{res}_h$.
The seasonal adjustment procedure is as follows.
First, the yearly and weekly seasonality is identified
by averaging the daily averaged data over 11 years
for each day of the year and for each weekday of the week, respectively.
The yearly seasonality is then smoothed because a single extreme value may overrepresent seasonality.
Second, the day-scale smoothed yearly and weekly seasonality
are subtracted from the hour-scale original data.
Finally, the daily seasonality and residuals are computed.
For the yearly seasonality, smoothing is done by locally weighted scatterplot smoothing \citet{Cleveland1979}.
\autoref{fig:seasonality} presents a sample result of seasonality adjustment of the target variables, PM\textsubscript{10} and PM\textsubscript{2.5}.
\autoref{fig:seasonality}\textbf{B, C, D} and \textbf{G, H, I} correspond to $s_{y,\textrm{smoothed}}$, $s_w$, and $s_h$, respectively,
whereas \autoref{fig:seasonality}\textbf{A, F} and \textbf{E, J} show the original value $x(t)$ and residual $r$.
The models used here, except for the OU process, accept input from residuals with standardization,
as mentioned in \autoref{subsec:data-collection}.

The most common way to determine stationarity is by examining the autocorrelation function (ACF),
which quantifies the linear interdependence between two instances of time series $x(t)$
separated by a discrete time lag $r (r > 0)$:

\begin{equation}
	\label{eq:acf}
	C(r) = \dfrac{\langle x(t) x(t+r) \rangle}{\langle x(t)^2 \rangle}
\end{equation}

If $x(t)$ and $x(t+r)$ are uncorrelated,
the correlation $C(r)$ becomes zero.
If two instances of a time series have a short-range correlation,
the dependence is characterized by rapid decay of the ACF, $C(s) \propto e^{-r/T}$,
where $T$ is the decay time or decorrelation time;
that is, $C(r)$ decreases to zero within a certain time lag $s^*$.

\begin{equation}
	\label{eq:short-acf}
	C(r) \propto e^{-r/T}
\end{equation}

For long-range dependence, the autocorrelation $C(r)$ is positive even for a large time lag $r$
and follows the power law

\begin{equation}
	\label{eq:long-acf}
	C(r) \propto r^{-\xi}\ \ \ 0 < \xi < 1
\end{equation}

If $\xi > 1$ the series is considered to have short-range dependence.
These facts depend on the convergence of the mean correlation distance,
$\bar{r} = \int^\infty_0 C(r) dr$, to a finite value.

\begin{figure}
	\centering
	\includegraphics[width=0.7\textwidth]{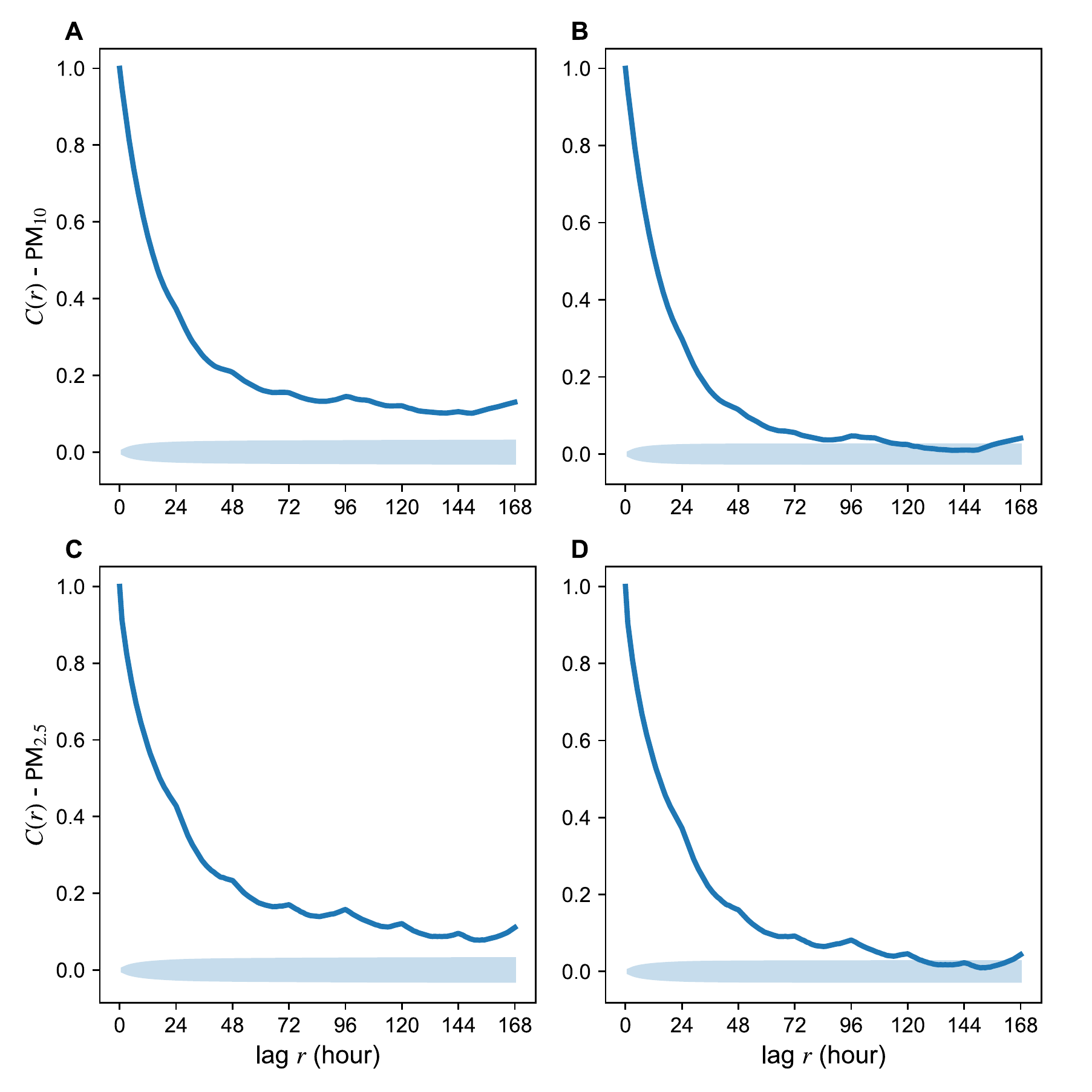}
	\caption{ACF of 7-day hourly data of PM\textsubscript{10} (upper) and PM\textsubscript{2.5} (lower).
		\textbf{A, C}:ACF without seasonal adjustment, \textbf{B, D}:ACF with seasonal adjustment.
		Shaded light blue: confidence interval of ACF.}
	\label{fig:acf}
\end{figure}

As shown in \autoref{fig:acf}, the ACF without seasonal adjustment shows a decay
to nonzero autocorrelation and an hourly seasonal pattern.
As mentioned in \autoref{sec:introduction}, environmental data have long-range dependence,
exhibiting not only a single seasonal pattern, but also multiple complex seasonal patterns.
PM\textsubscript{10}, PM\textsubscript{2.5} and the other variables
(except precipitation, which more closely resembles a sparse discrete event),
also have similar weekly and yearly patterns.
They were preprocessed according to the procedure for PM\textsubscript{10} and PM\textsubscript{2.5} before training.

After seasonal adjustment, the ACFs are closer to the confidence interval
than those without seasonal adjustment.
However, small-scale seasonal patterns remain.
Those patterns are trained using time2vec encoding, as discussed in \autoref{subsec:proposed}.

Quantitative analysis of the deseasonalized data was performed
using detrended fluctuation analysis (DFA) \citet{Kantelhardt2001,Koscielny-Bunde2006}.
DFA consists of four steps. First, the profile $Z(i)$ is defined by subtracting $\langle x \rangle$,
the mean of $x$, from $x_t$.

\begin{equation}
	\label{eq:dfa-profile}
	Z(i) \equiv \sum_{t=1}^i x_t - \langle x \rangle
\end{equation}

Second, the profile $Z(i)$ is divided into non-overlapping segments with size $s$.
Then, the detrended series $Z_s(i)$ of each segment $m$ is computed
by subtracting the local trends in the $m$th segment, $p_m (i)$,
from the original series $Z(i)$: $Z_s(i) \equiv Z(i) - p_m(i)$.
The local trend $p_m (i)$ is the fitting polynomial
for an $n$\textsuperscript{th}-order least-squares fit;
$n=2$ was used in this study.
Finally, the variance function from the detrended series $Z_s(i)$ is defined for the $m$th segment,
where $m=1,\dots,2N_s$.
In this study, quadratic polynomials ($n=2$) are used for fitting.

\begin{equation}
	\label{eq:dfa-variance}
	V_s^2(m) \equiv \dfrac{1}{s} \sum_{i=0}^{s} Z_s^2 [(m-1)s + i]
\end{equation}

and the mean fluctuation function is

\begin{equation}
	\label{eq:dfa-mean-fluc}
	V(s) = \left[ \dfrac{1}{2N_s} \sum_{m=1}^{2N_s} V_s^2 (m) \right]
\end{equation}

If a time series has long-range dependence, the DFA fluctuation function $V(s)$
increases for large segment size $s$ and follows another power law based on $\xi$ in \autoref{eq:long-acf},

\begin{equation}
	\label{eq:dfa-long}
	V(s) \sim s^h
\end{equation}

with

\begin{equation*}
	h = \begin{cases}
		1-\xi 	\; &\textrm{for} \; 0 < \xi < 1 \\
		1/2		\; &\textrm{for} \; \xi \geq 1
	\end{cases}
\end{equation*}

\begin{figure}
	\centering
	\includegraphics[width=0.7\textwidth]{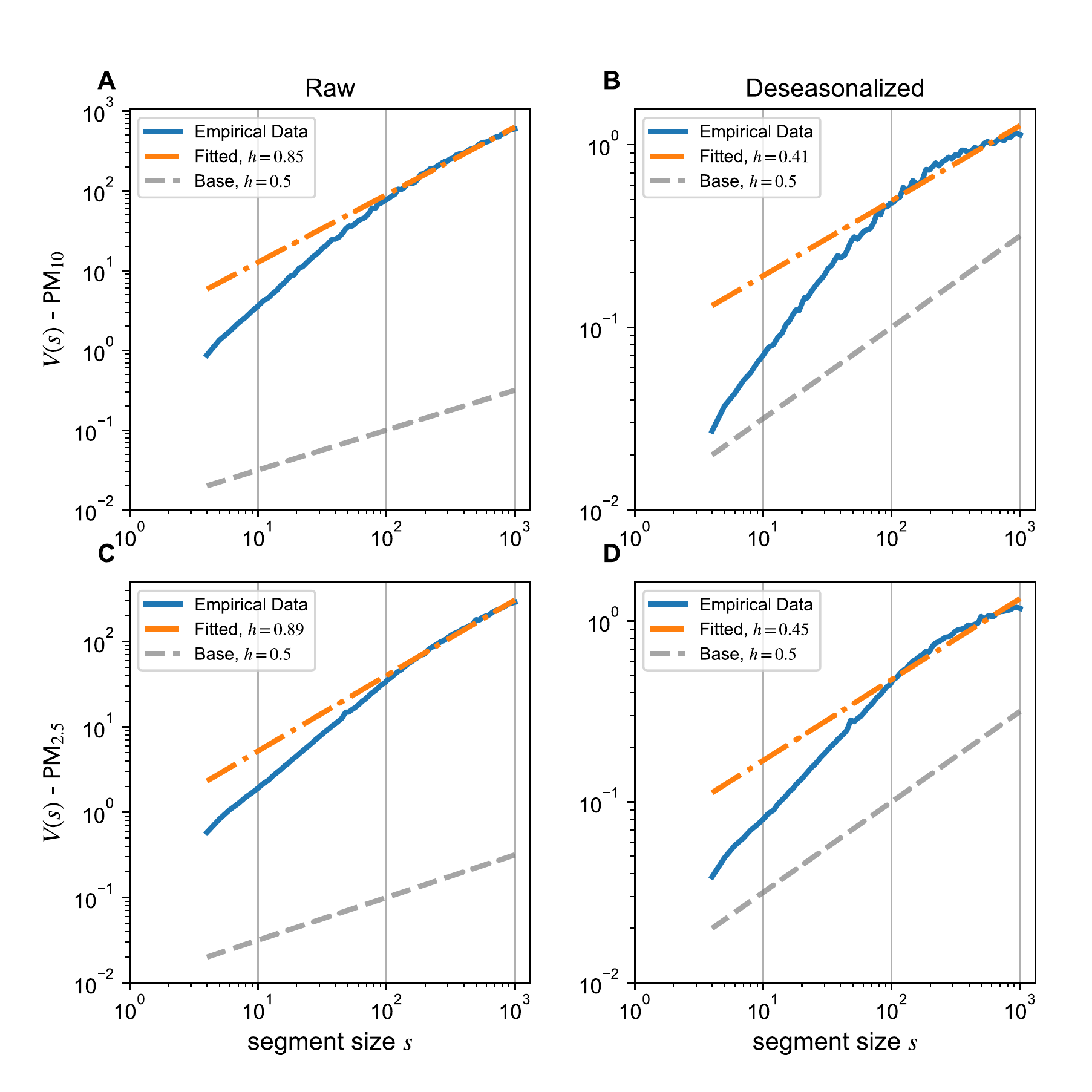}
	\caption{Mean fluctuation $[F(s)]$ and fitted line from DFA of
		PM\textsubscript{10} (above) and PM\textsubscript{2.5} (below).
		\textbf{A, C}: without seasonal adjustment, \textbf{B, D}: with seasonal adjustment.}
	\label{fig:dfa}
\end{figure}

In \autoref{fig:dfa}, the DFA fluctuation function is fitted using $s^h$ for large $s$.
Furthermore, the fluctuation exponent $h$ and the corresponding correlation exponent $\xi$
are presented in \autoref{tab:DFA}.
These two parameters approximately satisfy the condition given in \autoref{eq:dfa-mean-fluc}.
The $h$ values of the deseasonalized data (\autoref{fig:dfa}\textbf{B} and \textbf{D})
are approximately 0.5, and the $\xi$ are greater than 1, indicating short-range dependence.
In conclusion, the seasonal adjustment used in this study was quantitatively proven
to remove the long-range dependence of the raw data.

\begin{table}
	\caption{$h$ and $\xi$ values from DFA results.}
	\centering
	\begin{tabular}{clccc}
		\toprule
												& 								& $h$  & $\xi$ & $1-\xi$ 	\\ \midrule
		\multirow{2}{*}[-0.5ex]{PM\textsubscript{10}} 	& With Seasonality (\textbf{A}) & 0.85 & 0.31 	& 0.69		\\ \cmidrule{2-5}
												& With Seasonality (\textbf{B}) & 0.41 & 1.18 	& 			\\ \midrule
		\multirow{2}{*}[-0.5ex]{PM\textsubscript{2.5}} 	& With Seasonality (\textbf{C}) & 0.89 & 0.23 	& 0.77		\\ \cmidrule{2-5}
												& With Seasonality (\textbf{D}) & 0.44 & 1.11 	& 			\\
		\bottomrule
	\end{tabular}
	\label{tab:DFA}
\end{table}

\subsection{Heavy-Tailed Distribution}
\label{subsec:heavy-tailed}

The classical statistical modeling of time series usually describes noise as i.i.d. random variables.
However, in the real world, for example, in economics, computer science, and the environment,
noise has a heavy-tailed distribution, where the tail is heavier
than that of an exponential distribution.
A heavy-tailed distribution is problematic because the sampling errors are
large and negatively biased. When heavy-tailed data are used,
the error distribution of the model can be any distribution
except the normal distribution without affecting model performance.
It is clearly important to understand the data distribution before training a model.

\begin{figure}
	\centering
	\includegraphics[width=0.7\textwidth]{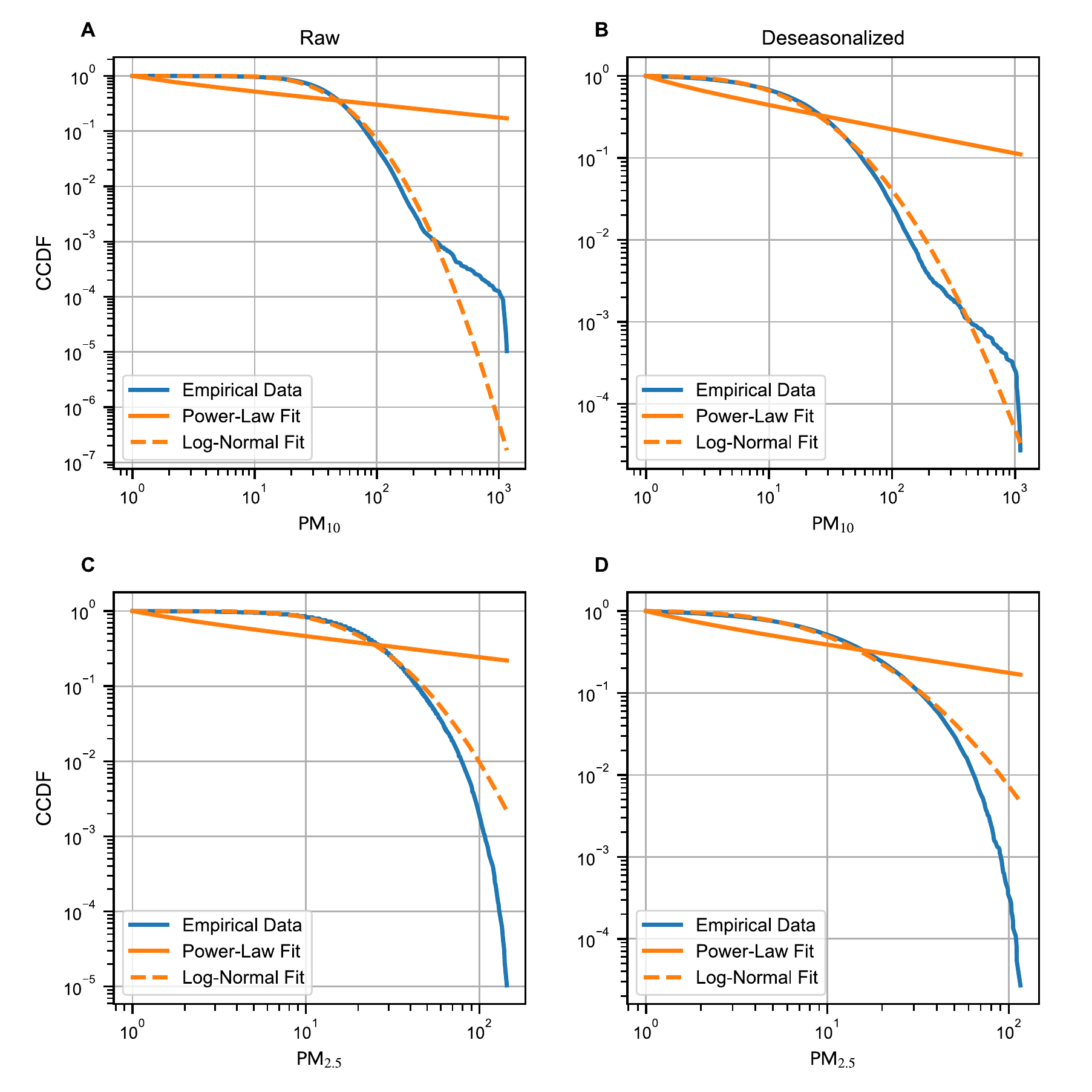}
	\caption{CCDF of PM\textsubscript{10}  and PM\textsubscript{2.5} data and fitted distributions.
		\textbf{A, C}: CCDF without seasonal adjustment,
		\textbf{B, D}: CCDF with seasonal adjustment.}
	\label{fig:ccdf}
\end{figure}

Although the most typical example of a heavy-tailed distribution
is the power law distribution (Pareto distribution),
the log-normal distribution should also be considered as an empirical data distribution \citet{Clauset2009}.
A complementary cumulative distribution function (CCDF) [$\bar{F}(x)=1-F(x)$,
where $F(x)$ is the cumulative distribution function] is generally used
to determine the distribution of heavy-tailed data.
The CCDF of the power-law distribution is given by the following equation.

\begin{equation}
	\label{eq:ccdf-pareto}
	\bar{F}(x) = Pr(X > x) = \begin{cases}
		\left( \dfrac{x_m}{x} \right)^\alpha 	\; &x \geq x_m  \\
		1										\; &x < x_m
	\end{cases}
\end{equation}

where $x_m$ is a scale parameter, and $\alpha$ is the shape parameter, or tail index.
The Pareto distribution is used to describe self-similar processes
owing to its scale invariance and the distribution of welfare in economics.
The log-normal distribution is a continuous probability distribution,
where the logarithm of random variables follows a normal distribution.
In a log-normal distribution, multiplication can be described
as taking the product of positive random variables.
The CCDF of the log-normal distribution is like that of the normal distribution,
where $\mu$ and $\sigma$ are the mean and standard deviation of distribution, respectively.

\begin{equation}
	\label{eq:ccdf-lognormal}
	\bar{F}(x) = \dfrac{1}{2} - \dfrac{1}{2} \textrm{erf}{\left[ \dfrac{\ln{x} - \mu}{\sqrt{2} \sigma} \right]}
\end{equation}

Power-law and log-normal distributions were fitted to the data with and without seasonal adjustment.
In \autoref{fig:ccdf}, the CCDF is plotted with positive data and fitted
by the log-normal and power-law distributions.
The powerlaw package (a Python package for analyzing heavy-tailed data distribution)
was used for the fitting \citet{Clauset2009,Alstott2014}.
This result shows that positive input data follow a log-normal distribution
rather than a power-law distribution.
Moreover, owing to the characteristics of the distribution,
the model accuracy clearly is not affected by seasonal adjustment.

This analysis clearly shows that the data used in this study
follow a heavy-tailed distribution but not a power-law distribution.
As shown in \autoref{fig:ccdf},
the distribution of the PM\textsubscript{10} data is better fitted
by the heavy tailed distribution than that of the PM\textsubscript{2.5} data.
To obtain better performance, the forecasting model should reflect the distribution.
Note also that the data used in the figure are filtered positive data,
although the seasonal adjustment introduces negative data naturally owing to subtraction.
Other regression methods based on the multiplication of random variables
cannot be applied owing to the negativity of the data and the nature of the logarithm.

\section{Method}
\label{sec:method}

Most of the methods used in this study are divided into two categories,
univariate and multivariate methods, according to the number of input parameters.
These methods are selected to compare the effect of the number of input variables
on the prediction results and address whether state-of-the-art methods are effective.
The methods are also categorized according to the methodological approach
as statistical models or machine learning models.
Two statistical models are considered: the Ornstein–Uhlenbeck (OU) process model,
which assumes that PM\textsubscript{10} and PM\textsubscript{2.5} are
stationary stochastic processes \citet{Uhlenbeck1930,Wojnowicz2012}, and the ARIMA model,
which is a state-space model \citet{Hyndman2018}.
The machine learning models are the XGBoost model,
which uses decision-tree-based ensemble machine learning \citet{Chen2016}; the MLP model,
a fully connected hierarchical neural network \citet{IanGoodfellowYoshuaBengio2015,Zhang1998};
an attention model based on a recurrent neural network (RNN) \citet{Hochreiter1997,Cho2014,Bahdanau2015};
and two more complex deep learning models,
the LSTNet and Time Series Transformer (TST) models \citet{Lai2018,Wu2020,Zerveas2020}.
The methods are summarized in \autoref{tab:methods}.

\begin{table}
	\begin{center}

	\caption{Methodologies}

	\label{tab:methods}
	\begin{tabular}{lll}
	\toprule
	Category & Type & Methods \\
	\midrule
	\multirow{4}{*}[-1.5ex]{Univariate} &  \multirow{2}{*}[-0.5ex]{Statistical} & Ornstein-Uhlenbeck process	   	\\ \cmidrule{3-3}
								&                               & ARIMA                         	\\ \cmidrule{2-3}
								&  \multirow{2}{*}[-0.5ex]{Machine Learning}  	& Multi Layer Perceptron 	\\ \cmidrule{3-3}
								&  										& Attention (Bahdanau) 		\\ \midrule
	\multirow{4}{*}[-1.5ex]{Multivariate}   & \multirow{4}{*}[-1.5ex]{Machine Learning} & XGBoost					\\ \cmidrule{3-3}
									& 									& Multi Layer Perceptron 	\\ \cmidrule{3-3}
									&									& LSTNet (Skip Layer) 		\\ \cmidrule{3-3}
									&                                	& Time Series Transformer   \\
	\bottomrule
	\end{tabular}
	\end{center}
\end{table}

\subsection{Univariate Models}
\label{subsec:univariate}

The univariate statistical methods used in this study are the Gaussian process model and ARIMA model.
The Gaussian process model is based on the OU process \citet{Uhlenbeck1930,Wojnowicz2012},
which evolves according to the following equation from the initial condition $x_0$:

\begin{equation}
	\label{eq:ou}
	dx=\dfrac{\mu - x}{\tau} dt + \sqrt{\dfrac{2\sigma^2}{\tau}}dW_t
\end{equation}

where $W_t$ denotes the standard Wiener process,
and $\mu$ and $\sigma$ are mean and standard deviation of the process, respectively.
The correlation time scale $\tau$ is computed
by integrating the ACF to the confidence interval cutoff in \autoref{fig:acf}.
The OU process is derived assuming that the time series is a simple stochastic Gauss–Markov process.
The following parameters are used to predict
the PM\textsubscript{10} and PM\textsubscript{2.5} concentrations:
mean $\mu$ values of -9.47 and -1.99, stationary variance $\sigma^2$ values of 458.39 and 215.15, and correlational time scales $\tau$ of 19.02 and 20.41, respectively.
The stochastic differential equation \autoref{eq:ou} was numerically solved using the Euler–Maruyama scheme \citet{Bayram2018}.

The ARIMA model is a widely used statistical model for time series analysis.
It is a linear combination of the AR model,
which used the number of lags of dependent variables;
the MA model, which uses the number of lags of the forecast errors;
and an integrated component, which uses differenced time series to obtain stationary time series.
It is a generalized version of the autoregressive moving average model.
As described in \autoref{subsec:seasonal-adjustment},
the time series data used in this study become stationary;
therefore, the integration component can be ignored.
The strategy for determining the order of the AR and MA terms using the ACF
and partial autocorrelation function is described in \citet{Hyndman2018}.
The resulting equation for the AR($p$) model, where $p$ is the order of the autoregressive term, is

\begin{equation}
	\label{eq:ar}
	Y_t = c + \phi_1 Y_{t-1} + \cdots + \phi_p Y_{t-p}
\end{equation}

The coefficients and constant of the AR model, $\phi_i$ and $c$,
are estimated using conditional and exact maximum likelihood and conditional least-squares,
respectively, as implemented in stats-models (a Python package for statistical models)
using the training set data \citet{Seabold2010}.

The MLP and RNN-based attention mechanism models are used as univariate machine learning models.
The MLP model is a general class of feedforward artificial neural network.
MLP models show promising results for forecasting because they
learns adaptively to be capable of flexible nonlinear modeling \citet{Zhang1998}.
In an MLP model, all input data are serialized in one layer,
and hidden layers are distributed in multiple layers.
Each node receives an input signal as information,
transforms it by an activation function, processes it by multiplying the local weight of each node,
and then passes it to the output. L
earning and changing the weights in the perceptron is performed
after all the data are processed using the error of the output through back-propagation.
Leaky ReLU is used as the activation function.

A disadvantage of MLP models is that they cannot handle the temporal information of the data
because each node processes data independently.
The RNN is designed for problems involving temporal sequences, such as machine translation.
RNN models have an internal state to process input data depending on its shape
and the dimensions of the model.
Because an RNN uses repeated multiplication of the recurrent weight matrix,
it has an exploding/vanishing gradient problem and
cannot easily access very old information.
LSTM and the gated recurrent unit (GRU) have been developed
to resolve this problem by adding multiple gates in the internal state
to store long-term information \citet{Hochreiter1997,Cho2014}.
At each LSTM or GRU cell, they learn what to learn and what to forget from a single input.
Their output is called the hid-den state from last cell,
and it can be fed into the next cell recursively.

\begin{figure}
	\centering
	\includegraphics[clip,trim=0cm 2.5cm 0cm 0cm,width=0.9\textwidth]{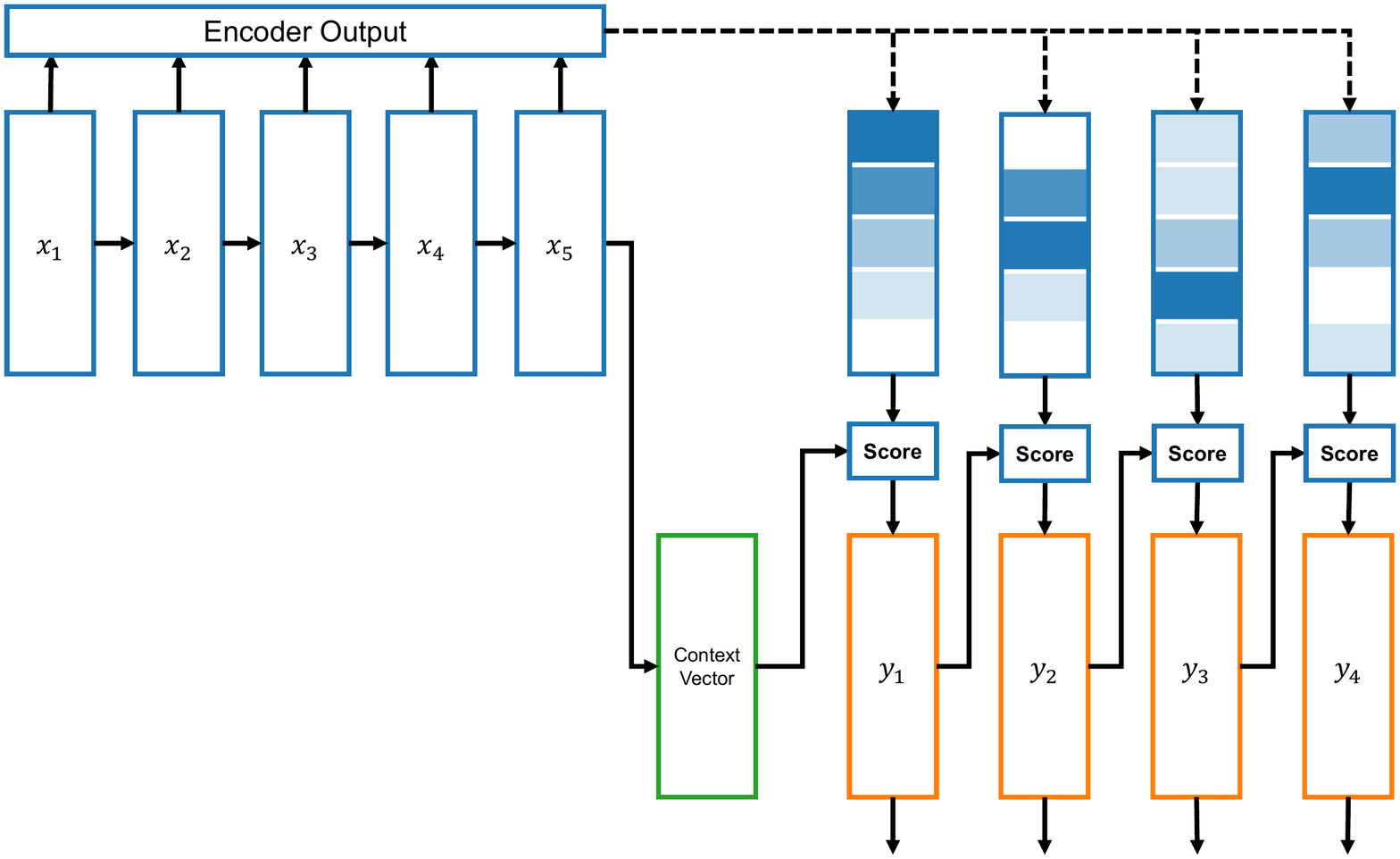}
	\caption{Schematic diagram of attention mechanism. Each output sequence $\{y_1, \cdots, y_4 \}$
		is generated by a hidden state starting from a single context vector from input sequence
		$\{x_1, \cdots, x_5\}$ and an attention vector. The blue-shaded vectors indicate
		how much the input sequence "attends" to the output seqeunce}
	\label{fig:attention}
\end{figure}

Another improvement in machine translation resulting from the improved single-node architecture
mentioned above is the attention mechanism \citet{Bahdanau2015}.
This mechanism is based on a sequence-to-sequence (S2S) technique,
which translates input to output by an encoder-decoder paradigm \citet{Sutskever2014}.
The encoder consumes input data and compresses them with a single latent vector;
then the decoder learns to produce output data from the latent vector.
This technique makes input-output translation possible in domain-to-domain mapping rather than one-to-one.
Although the S2S technique translates domain-to-domain mapping,
the encoder compresses the whole input data into a single latent vector.
Consequently, some information may be lost during compression.
In addition, the decoder also has a problem; it treats all input sequences
as having the same importance because its input is the single latent vector
generated by the encoder.
The importance of each node in the sequence may vary, and this variation must be assessed.
The attention mechanism “attends” to the input sequence when the decoder generates data.
As shown in \autoref{fig:attention}, the alignment score is calculated
using a combination of the output of each encoder,
the decoder input (context vector), or the last hidden state at each decoder step.
The alignment score quantifies how the model attends to its input
by choosing multiple methods such as concatenation or taking the dot product.
In this study, a GRU based on the concatenation score method was chosen
to decrease the learning time and increase the accuracy.

\subsection{Multivariate Models}
\label{subsec:multivariate}

Since the environmental variables depend on multiple factors simultaneously,
a univariate model may not be sufficient for PM\textsubscript{10} and PM\textsubscript{2.5} forecasting.
Multivariate models are suggested to resolve the problem.
In multivariate models, it is important to capture nonlinear patterns
between variables even when a variable itself exhibits a temporal pattern.

The multivariate MLP model architecture is identical to that of a univariate model,
but the input data are serialized as one-dimensional.
XGBoost is a fast and performant gradient boosting decision-tree model.
A decision tree model constructs a probability-based tree,
and prediction is performed by proceeding from a node (branch) to the target (leaves).
A single decision tree is vulnerable to overfitting
due to a single path to a leaf and is known as a weak learner that is only slightly
better than random chance.
XGBoost is an ensemble model using a gradient boosting strategy with CART,
and it combines multiple decision trees with weight and loss function optimization \citet{Chen2016}.
It has rapidly become popular because it is faster than existing gradient-boosting implementations
and is memory efficient.

LSTNet is a multivariate time series model combining a convolutional component
with a convolutional neural network (CNN) without pooling
and a recurrent component with a GRU \citet{Lai2018}.
In the convolutional component, the CNN is used to extract patterns
involving multiple features and short-term temporal patterns.
The result of the convolutional component is processed
by the recurrent component to capture longer-term patterns.
In LSTNet, recurrent-skip and autoregressive components are also added
to memorize periodic patterns and the scale of the input.
Although LSTNet outperforms other models on bench-mark datasets
(traffic, solar energy, and electricity datasets),
which show highly periodic patterns even in input length,
but not on an exchange rate dataset, which shows only nonperiodic patterns.
This result indicates that more complex models are needed.

The attention mechanism allows a model to attend
to a specific part of the input when producing output.
Although it yields better machine translation results
than typical RNN models because it enables the translation of longer sequences,
the importance of differences in attention between input sequences is increased
when meaning is determined from the input itself, for example,
in the translation of pronouns in a sentence.
Self-attention, or intra-attention, is used to extract different features within an input sequence.
Furthermore, RNN-based models are vulnerable to problems with longer sequences because of the
exploding/vanishing gradient problem, and they cannot be parallelized.
Therefore, a self-attention-based transformer model was introduced to use attention
instead of a recurrent unit \citet{Vaswani2017}.
The transformer outperforms existing models; moreover,
it enables the parallelization of the model because it uses feed-forward layers
instead of a recurrent network unit, which can only be processed in serial order.
Owing to this success, the transformer has also been applied in time series forecasting
and showed promising results \citet{Wu2020,Zerveas2020}.
The TST, an encoder-based transformer architecture \cite{Zerveas2020},
is used in this study because of its simplicity and performance.
Even though the TST was not designed for forecasting problems,
its encoder architecture is applicable to these problems.
Here, we improved the TST model, as explained in the next section.

\subsection{Proposed Models}
\label{subsec:proposed}

When the given time series window is $\mathbf{X_T}=\{\mathbf{x_1}, \mathbf{x_2}, \cdots, \mathbf{x_T}\}$,
where $\mathbf{x_t} \in \mathbb{R}^n$, and $n$ is the number of variables,
the forecasting task for output horizon $h$ is to predict
$\mathbf{Y_T}=\{\mathbf{y_{T+1}}, \mathbf{y_{T+2}}, \cdots, \mathbf{y_{T+h}}\}$.
In this study, $\mathbf{Y_T}$ is a single variable,
PM\textsubscript{10} or PM\textsubscript{2.5}, and the output matrix is $\mathbf{Y_T}\in\mathbb{R}^n$,
where $h$ is the output horizon size, and the input matrix is $\mathbf{X}\in\mathbb{R^{n\times T}}$.

Most machine-learning-based models are specialized to capture nonlinear patterns in time series.
We also added a learnable autoregressive component with a single fully connected layer,
as in the LSTNet model.

\begin{equation}
	\label{eq:lin-nonlin}
	\hat{Y}_t = \mathrm{AR}(\mathbf{X^{\mathrm{1d}}_T}) + \mathrm{model} (\mathbf{X_T})
\end{equation}

where $X^{\mathrm{1d}}_T$ is one-dimensional target variable,
and $\mathbf{X_T}$ is a multivariate whole sliding window input.

The overall architecture of the proposed model of the encoder-only transformer
is the same as that in \citet{Zerveas2020}.
The positional encoding, however, is replaced with Time2Vec \citet{Kazemi2019},
a learnable parametric positional encoding using a sine function,
because periodic patterns remain, as shown in \autoref{fig:acf}.
Time2Vec consists of the following two components.

\begin{equation}
	\label{eq:time2vec}
	\mathbf{t2v}(\tau)[i] = \begin{cases}
		\omega_i \tau + \phi_i				\ &\textrm{if}\ \ i=0 \\
		\mathcal{F}(\omega_i \tau + \phi_i)	\ &\textrm{if}\ \ 1 \leq i \leq k
	\end{cases}
\end{equation}

where $\mathbf{t2v}(\tau)[i]$ is the $i$\textsuperscript{th} element of $\mathbf{t2v}$,
$\mathcal{F}$ is a periodic activation function (sine function),
and $\omega_i$ and $\phi_i$ are learnable parameters
such as the period and phase shift in a periodic function.
If the length of $\mathbf{t2v}$ is 1, \textemdash $i=0$,
it is the same as the learnable positional encoding proposed in \citet{Zerveas2020}.
The improved architecture is illustrated in Fig. 6.

\begin{figure}
	\centering
	\includegraphics[clip,trim=0cm 2.5cm 0cm 0cm,width=0.9\textwidth]{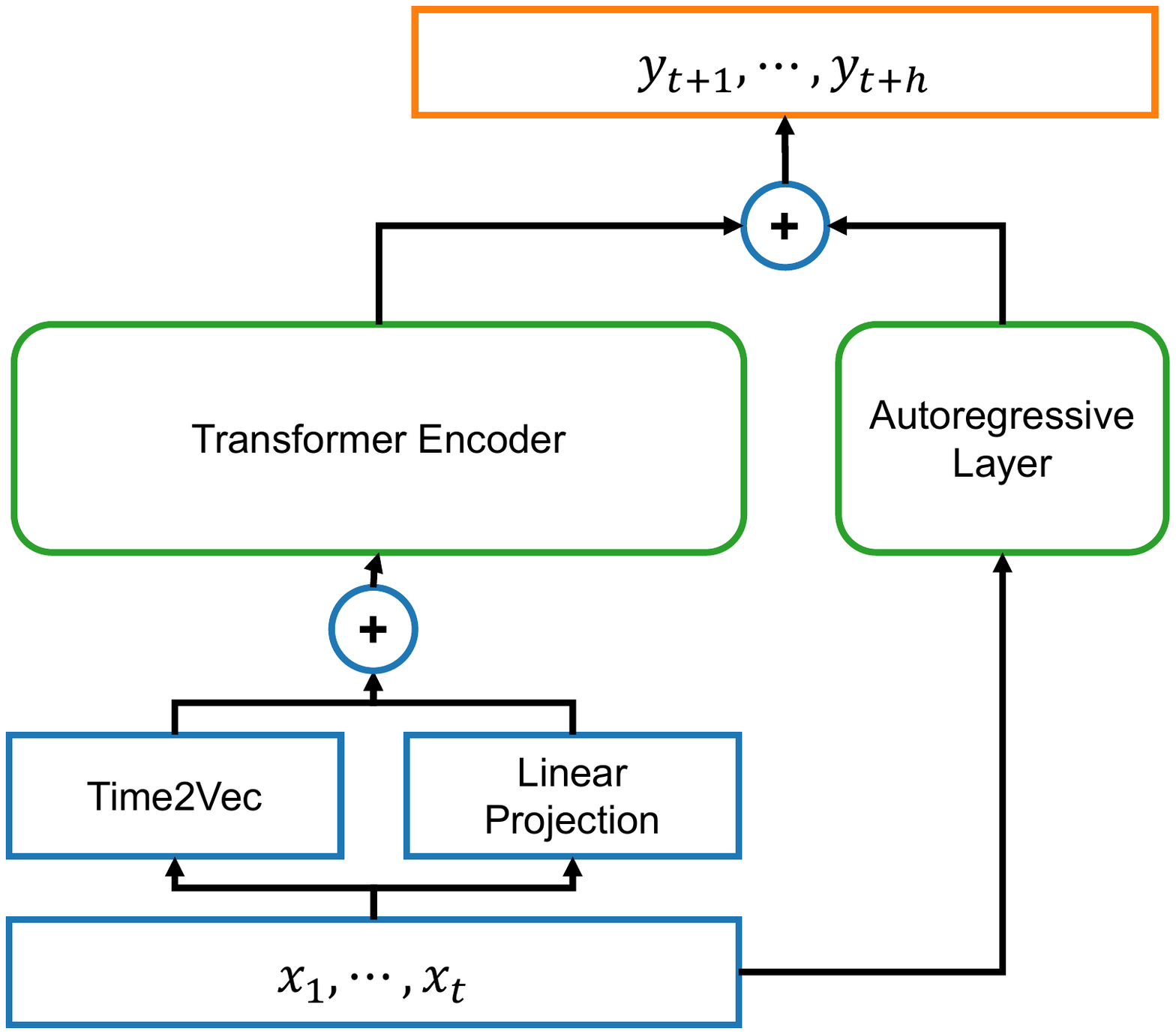}
	\caption{Schematic diagram of modified TST}
	\label{fig:TST}
\end{figure}

It has been argued \citet{Zerveas2020} that batch normalization
between every self-attention layer and feed-forward layer
results in better performance than default layer normalization \citet{Vaswani2017}.
However, a model with batch normalization was unstable and unsuitable for our data;
thus, we applied layer normalization instead.
The code used in this study is publicly available at GitHub (\url{https://github.com/appleparan/mise.py}).

\subsection{Maximum Correntropy Criterion Induced Losses for Regression}
\label{subsec:mccr}

The loss function is an objective function between the observed and predicted values.
Most deep learning algorithms involve optimization problems
in which a certain loss function is minimized or maximized.
The most common loss function used in regression problems is the mean squared error (MSE) loss.

\begin{equation}
	\label{eq:mse}
	\textrm{MSE} (Y_t, \hat{Y}_t) = \dfrac{1}{n} \sum_{i=1}^n (Y_i - \hat{Y}_i)^2
\end{equation}

The MSE loss, however, tends to suffer from extreme values,
which is common for heavy-tailed distributions because the error is squared.
Although numerous efforts have been made to predict extreme values
(for example, the use of a memory network module and a variable loss function \citet{Ding2019,Ribeiro2020}),
these methods require a long training time
because they need to access a different dataset for each batch.

It is necessary to consider the data distribution
when training models based on a heavy-tailed distribution.
The Kullback–Leibler (KL) divergence, $D_{KL} (P \mid\mid Q) = \sum_x P(x) \log{P(x)/Q(x)}$,
is commonly used to compare the distributions of random variables $P$ and $Q$.
In \citet{Qi2020}, the KL divergence is used as a loss function
to predict extreme values in terms of the difference
between the probability distributions in complex systems, such as turbulence modeling.
It is simple to implement and can be trained more rapidly than previous functions,
but as the authors of \citet{Qi2020} mentioned,
they are interested in statistical features rather than the exact trajectory of the system,
and small shifts in extreme values are negligible for their purpose.

\begin{figure}
	\centering
		\includegraphics[width=0.5\textwidth]{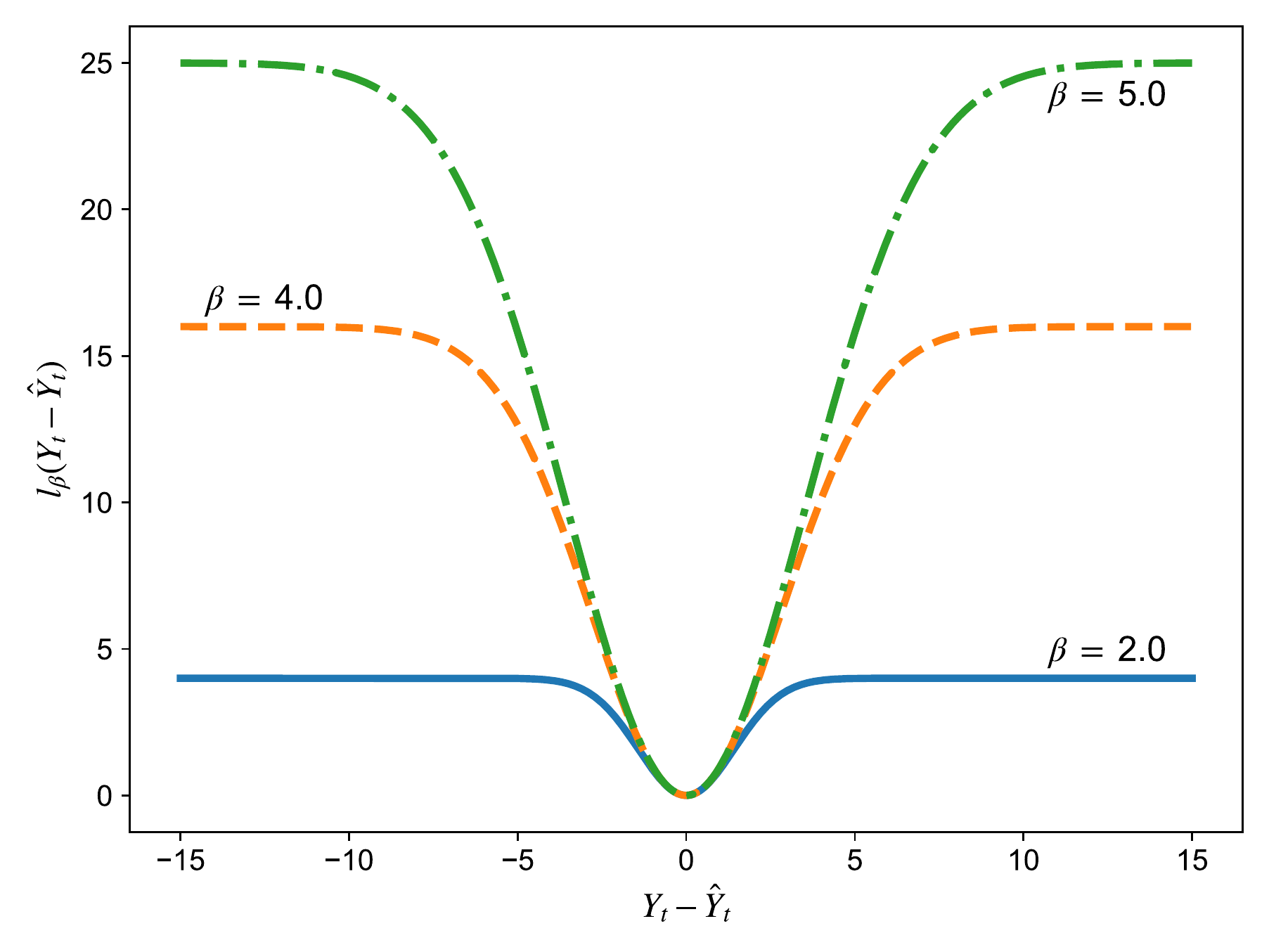}
	\caption{Actual and predicted values of MCCR for different $\beta$
		corresponding to $Y_t - \hat{Y}_t$}
	\label{fig:MCCR}
\end{figure}

To overcome the limitations of the KL divergence,
the correntropy has been proposed as a robust localized similarity measure \citet{Liu2007}.
The correntropy generalizes the correlation by nonlinear mapping of random variables
and measures the similarity between random variables by information theoretic learning.
The maximum correntropy criterion has been used as a loss function
and applied to regression problems \citet{Liu2007,Feng2015}.
The maximum correntropy criterion based regression (MCCR) loss is denoted as

\begin{equation}
	\label{eq:mccr}
	l_\beta (Y_t, \hat{Y}_t) = \beta^2 (1 - e^{-(Y_t - \hat{Y}_t)^2/\beta^2})
\end{equation}

where $\beta$ is a positive scale parameter, and $Y_t$ and $\hat{Y}_t$
are the actual and predicted values, respectively.
MCCR is plotted for various $\beta$ values in \autoref{fig:MCCR}.
As suggested by \citet[Theorem 8]{Feng2015}, the MCCR model and least square model are equivalent
if a sufficiently large $\beta$ is chosen.
The authors of \citet{Feng2015} also proved that the MCCR can model heavy-tailed noise,
whereas the conventional least square model handles only sub-Gaussian noise.
For this reason, and owing to the data distribution used in this study,
the MCCR loss is used here, and the results are compared with those of the MSE loss.

As noted in the explanation of the OU process, the confidence interval cutoff,
or correlation timescale, of PM\textsubscript{10} and PM\textsubscript{2.5} is approximately 20.
For training a small periodic pattern using $\mathbf{t2v}$,
the minimum input size should be at least 48; thus, this value was chosen.

\section{Results}
\label{sec:results}

\begin{figure}
	\centering
		\includegraphics[width=0.5\textwidth]{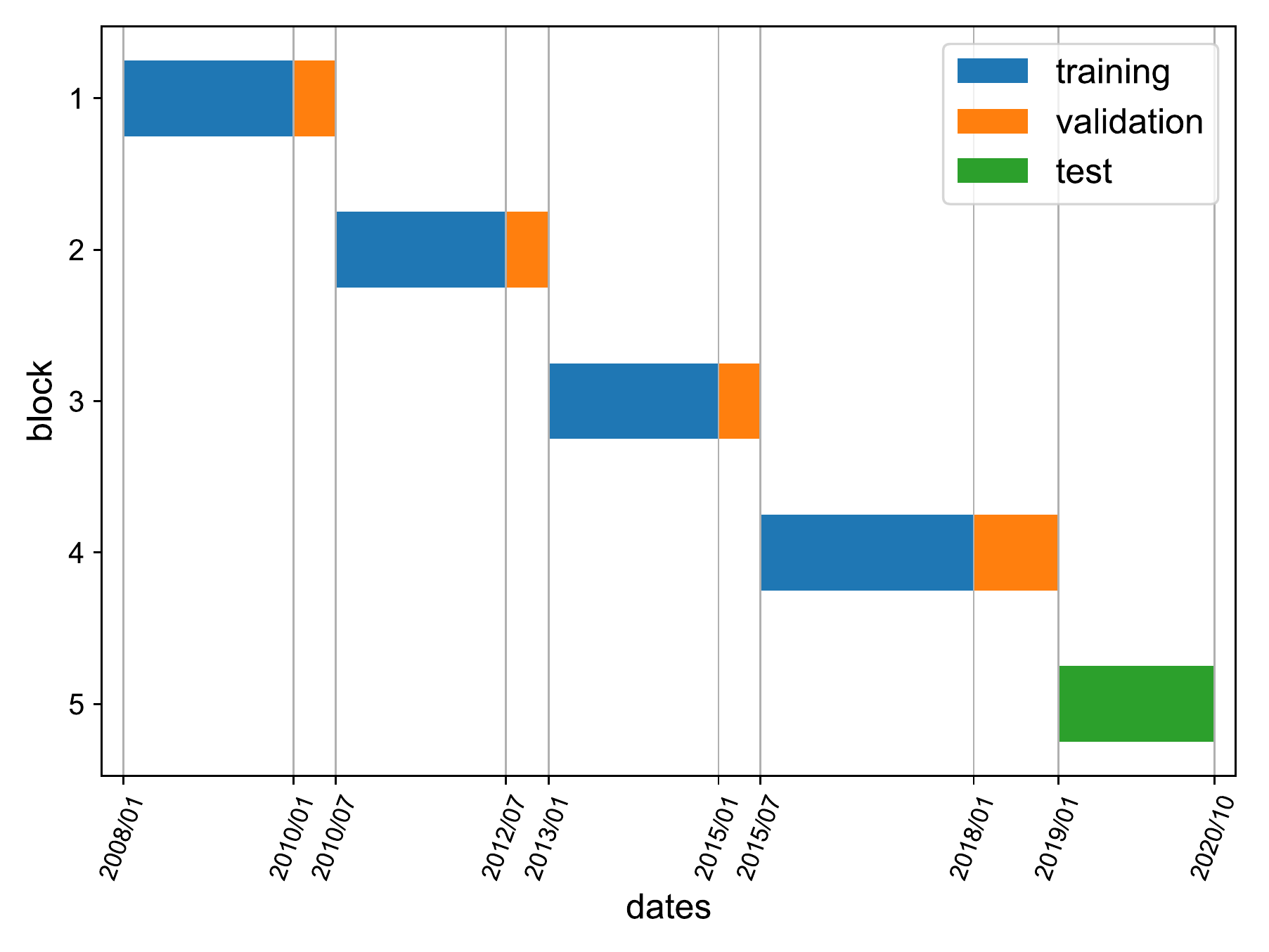}
	\caption{Diagram of blocked cross-validation region for training, validation, and test sets.}
	\label{fig:CV}
\end{figure}

Cross-validation is another important topic in time series forecasting
because of the dependence of the sliding window on the time series.
Typical cross-validation splitting methods
such as random splitting may cause overfitting between the training, validation, and test sets.
To avoid this problem, blocked cross-validation was used, as shown in \autoref{fig:CV}.
The training and validation sets are mixtures of multiple seasons
for proper training of the seasonal variation.
The ratio between the training, validation, and test sets is approximately 67\%/20\%/13\%.

For multistep horizon prediction, a multiple-input multiple-output strategy is applied
to most machine learning models, except for the OU process and XGBoost. Because the OU process
is naturally unable to produce multiple outputs at once,
a recursive strategy of iterative forecasting was applied.
XGBoost is also unable to consume a multivariate sliding window
and produce multiple outputs; thus, a single-Input single-output strategy was applied
by creating multiple models for each horizon length
and mapping the last step of the input to each model.

\subsection{Hyperparameter Optimization}
\label{subsec:optimization}

For hyperparameter optimization, Bayesian optimization was used in the form
of the Tree-structured Parzen Estimator sampler implemented
in Optuna (a hyperparameter tuning framework) \citet{Akiba2019}.
\autoref{tab:optimized-alpha} lists the optimized scaling parameters of the MCCR loss.
Some parameters are excluded from the optimization,
such as the sliding window input length, output horizon size, and batch size,
which are 48, 24, and 64 h, respectively.
For weight and bias optimization, Adam was used with a learning rate of $10^{-4}$ \citet{Kingma2015}.
Ridge regularization (L2), the most popular regularization technique for time series forecasting,
was used. PyTorch is used to train the deep learning networks \citet{Paszke2019}.

\begin{table}
	\begin{center}

	\caption{Optimized scaling parameter $\alpha$ of MCCR loss}

	\label{tab:optimized-alpha}
	\begin{tabular}{llll}
	\toprule
	\multirow{5}{*}[-2.5ex]{PM\textsubscript{10}} 	&  \multirow{2}{*}{Univariate} & Multi Layer Perceptron & 4.05	\\ \cmidrule{3-4}
											&                              & Attention & 4.45 \\ \cmidrule{2-4}
											&  \multirow{3}{*}[-1ex]{Mutlivariate}	& Multi Layer Perceptron & 4.60 	\\ \cmidrule{3-4}
											&  									& LSTNet (Skip) & 2.60 \\ \cmidrule{3-4}
											&  									& Time Series Transformer & 4.10 \\
	\midrule
	\multirow{5}{*}[-2.5ex]{PM\textsubscript{2.5}} 	&  \multirow{2}{*}{Univariate} & Multi Layer Perceptron & 4.65	\\ \cmidrule{3-4}
											&                              & Attention & 4.15 \\ \cmidrule{2-4}
											&  \multirow{3}{*}[-1ex]{Mutlivariate}	& Multi Layer Perceptron & 7.60 	\\ \cmidrule{3-4}
											&  									& LSTNet (Skip) & 2.10 \\ \cmidrule{3-4}
											&  									& Time Series Transformer & 4.10 \\
	\bottomrule
	\end{tabular}
	\end{center}
\end{table}

Optuna selects hyperparameter by sampler and determines
whether it is worth to run trial by running single epoch.
If not, Optuna drops the selected hyperparameter and goes forward to next trial,
which is called pruning.
Because of pruning, hyperparameter optimization time varies by case.
Some initial trials have fixed parameters with uniform intervals
to avoid falling local minimum and each trial runs 20 epochs.
After optimizing hyperparameters,
our models are trained by best hyperparameter set up to 500 epochs.
TST model have 160 maximum trials which is the largest among our models.
Usually, TST model took a day for optimization, 4 hours for training,
few minutes for testing.
The models are trained and tested using NVIDA TITAN Xp GPU and Intel Xeon Gold 6140 CPU.

The deterministic models depend on other models output
such as weather and emission models,
which also require extensive computing resources.
In \citet{Myoung2018},
they indicate that national air quality forecasting system runs
weather, emission, and air quality models and takes 3 hour 40 minutes
even without preprocessing and postprocessing
while our models need few minutes for forecasting with trained weights.
In South Korea, air quality forecasting is done by every 6 hours
and models should run on every prediction.
Statistical models or machine learning models reuse trained parameters
thus they can save a lot of computing resources.

\subsection{Evaluation Metrics}
\label{subsec:metrics}

It is important to choose the best metric for evaluating models.
The most widely used evaluation metrics for regression problems are the MSE,
mean absolute error, and correlation coefficient (CORR).

Despite the simplicity and intuitive interpretation of those metrics,
they are overly influenced by even a few extremely large values.
Relative difference of evaluation metrics,
such as the normalized MSE and normalized MAE are alternatives,
but they are historically normalized by observed values;
consequently, the metric is inflated if the observations are small values.
Several metrics have been proposed for intuitive, symmetric,
and unbiased model performance evaluation \citet{Yu2006},
such as the normalized mean bias factor (NMBF) and normalized mean absolute error factor (NMAEF).

\begin{equation}
	\label{eq:NMBF}
	\textrm{NMBF} = \begin{cases}
		\dfrac{\bar{M}}{\bar{O}} - 1	&\ \textrm{if}\ \bar{M} \geq \bar{O} \\
		1 - \dfrac{\bar{O}}{\bar{M}}	&\ \textrm{if}\ \bar{M} < \bar{O} \\
	\end{cases}
\end{equation}

\begin{equation}
	\label{eq:NMAEF}
	\textrm{NMAEF} = \begin{cases}
		\dfrac{\sum{|M_i - O_i|}}{\sum{O_i}}	&\ \textrm{if}\ \bar{M} \geq \bar{O} \\[10pt]
		\dfrac{\sum{|M_i - O_i|}}{\sum{M_i}}	&\ \textrm{if}\ \bar{M} < \bar{O} \\
	\end{cases}
\end{equation}

where $\bar{M}, \bar{O}$ is the mean
of the model prediction (predicted, $\hat{Y}_t$ ) and observation (actual, $Y_t$ ).
These metrics are symmetric and unbiased because their values change
according to the ratio of the observed and modeled quantity.
The sign of the NMBF indicates overestimation and underestimation,
where a positive NMBF indicates overestimation,
whereas a negative NMBF indicates underestimation by a factor.
The NMAEF is considered a robust normalized error metric
for extreme values because the denominator of the NMAEF differs by
the ratio of the observed and modeled quantities.
In this study, in addition to the NMBF and NMAEF, conventional metrics, specifically,
the root mean squared error (RMSE) for readability and CORR are also used.
A smaller NMAEF indicates better results because it is a type of error;
because the NMBF indicates bias, a value closer to 0 is better.
The RMSE, like the MSE, indicates better results when the value is smaller,
whereas a higher CORR is better.

\begin{equation}
	\label{eq:RMSE}
	\textrm{RMSE} =
		\dfrac{1}{N} \sqrt{\sum_{i=0}^N (M_i - O_i)^2}
\end{equation}

\begin{equation}
	\label{eq:CORR}
	\textrm{CORR} =
		\dfrac{1}{N} \dfrac{\sum_i (M_i - \bar{M}) (O - \bar{O})} {\sqrt{\sum_i (M_i - \bar{M})^2 \sum_i (O_i-\bar{O})^2}}
\end{equation}

\subsection{Performance Results}
\label{subsec:performance}

In model performance evaluation, it is important to understand the prediction error.
The risk, that is, the average loss across all the data,
can be decomposed into the reducible error and irreducible error,
which is the lowest bound of the generalization error.
Moreover, the reducible error is decomposed into bias and variance.
As explained in \autoref{subsec:metrics}, the NMBF measures the bias of the results,
and the NMAEF indicates variance, whereas the MSE,
$\textrm{MSE}(\hat{Y}) = E[(\hat{Y} - Y)^2]=\textrm{Var}(\hat{Y})+\textrm{Bias}(\hat{Y},Y)^2$,
measures the sum of variance and bias.
For noisy data such as air pollution data,
the positive irreducible error $\epsilon$ is also represented in the MSE.
In this section, we select the best models and analyze the errors
by decomposing the bias and variance of each model.
In addition, the performance of the loss functions is compared
to provide insights into the selection of an appropriate loss function
according to the data distribution.

\begin{figure}[!htb]
	\centering

	\includegraphics[width=\textwidth]{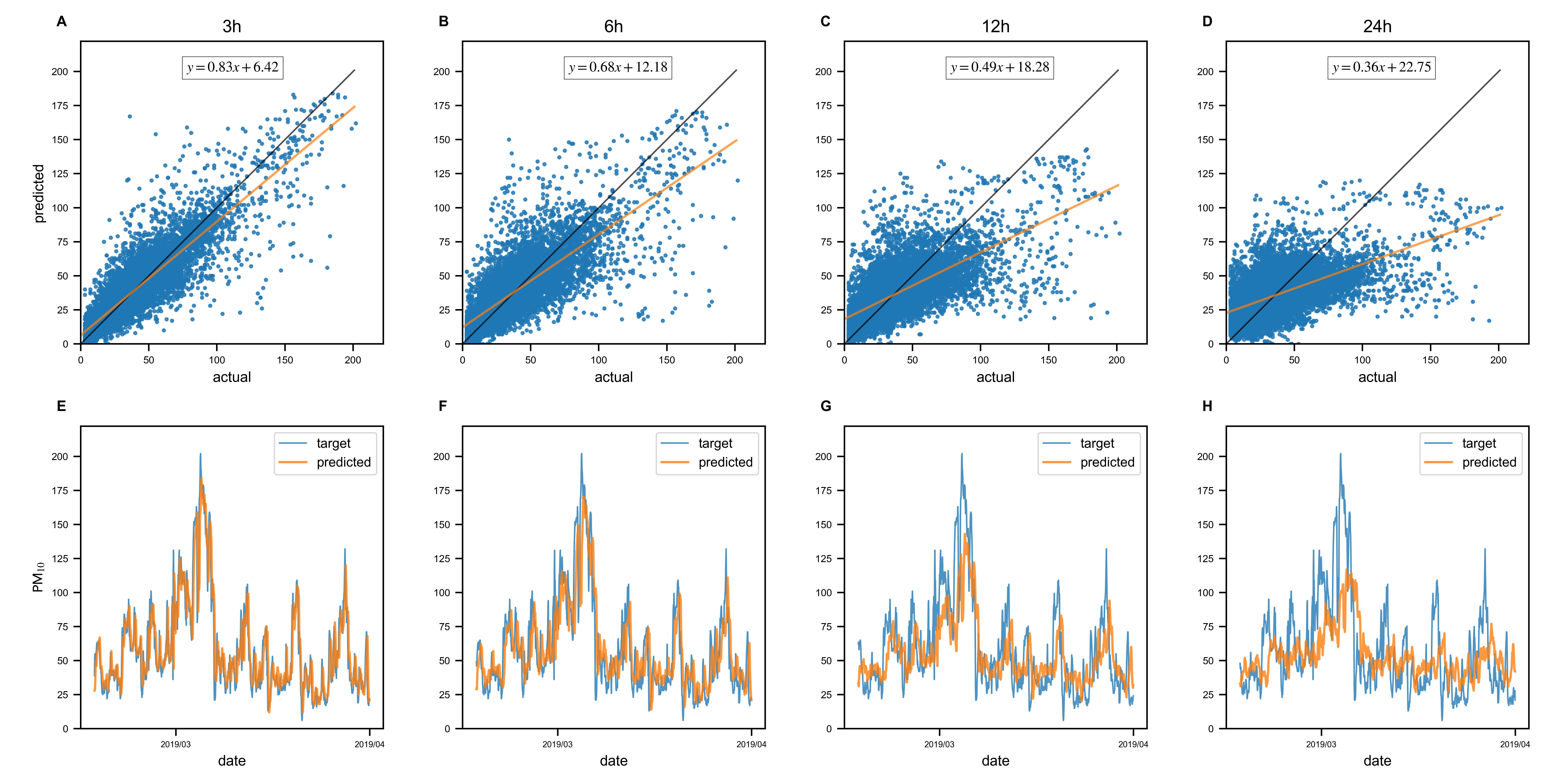}
	\caption{Scatter and line plot of MCCR of PM\textsubscript{10} TST model results for various horizons.}
	\label{fig:horizon-PM10}

	\includegraphics[width=\textwidth]{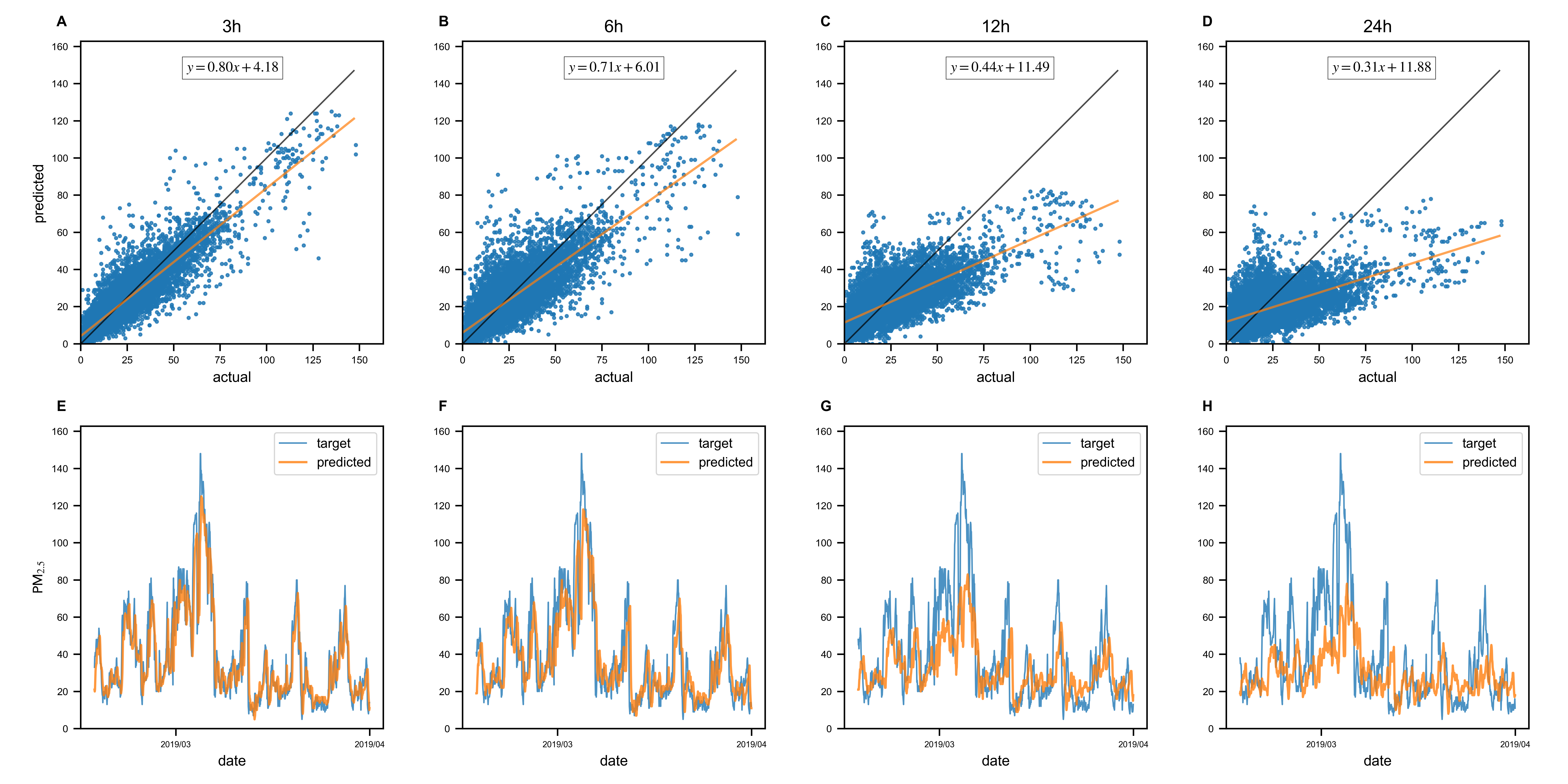}
	\caption{Scatter and line plot of MCCR of PM\textsubscript{2.5} TST model results for various horizons.}
	\label{fig:horizon-PM25}

\end{figure}

Three main factors are compared in this study: the horizon, model, and loss function.
We compare them in this order and then generalize the results.
In \autoref{fig:horizon-PM10} and \autoref{fig:horizon-PM25},
PM\textsubscript{10} and PM\textsubscript{2.5} model prediction
results for various horizons are compared.
For shorter horizons, the model shows high accuracy, as expected.
However, for longer horizons, the model accuracy decreases because the model error accumulates;
in addition, high peaks are not predicted.
This tendency is clear when the horizon is longer.
Thus, the 24 h horizon, the longest horizon considered here,
was used for further analysis to ensure rigorous assessment.

\begin{figure}[!htb]
	\centering

	\includegraphics[width=\textwidth]{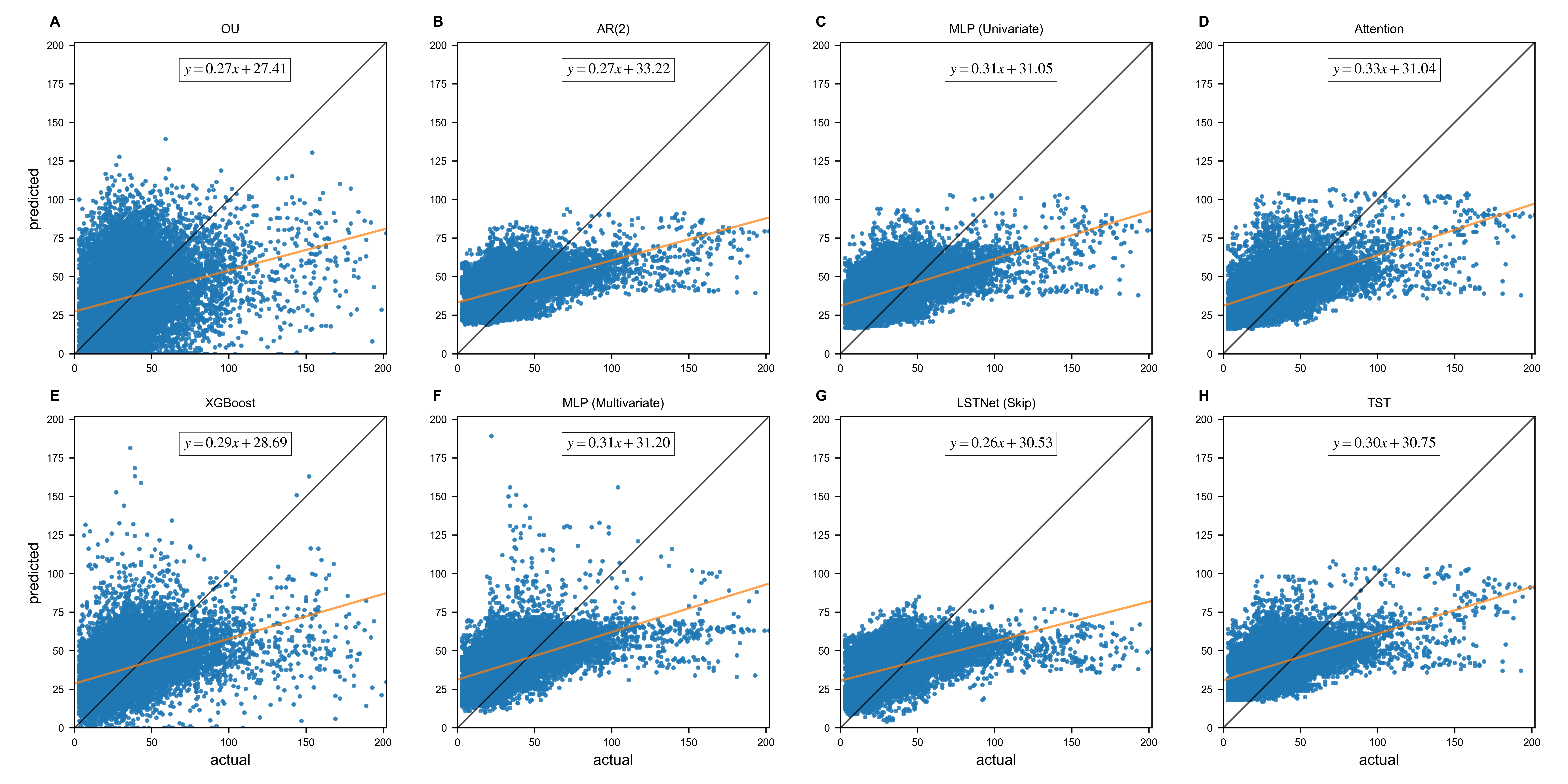}
	\caption{Scatter plot of the result of PM\textsubscript{10} models using MSE loss. Equations give the best fit line.}
	\label{fig:scatter-mse-PM10}

	\includegraphics[width=\textwidth]{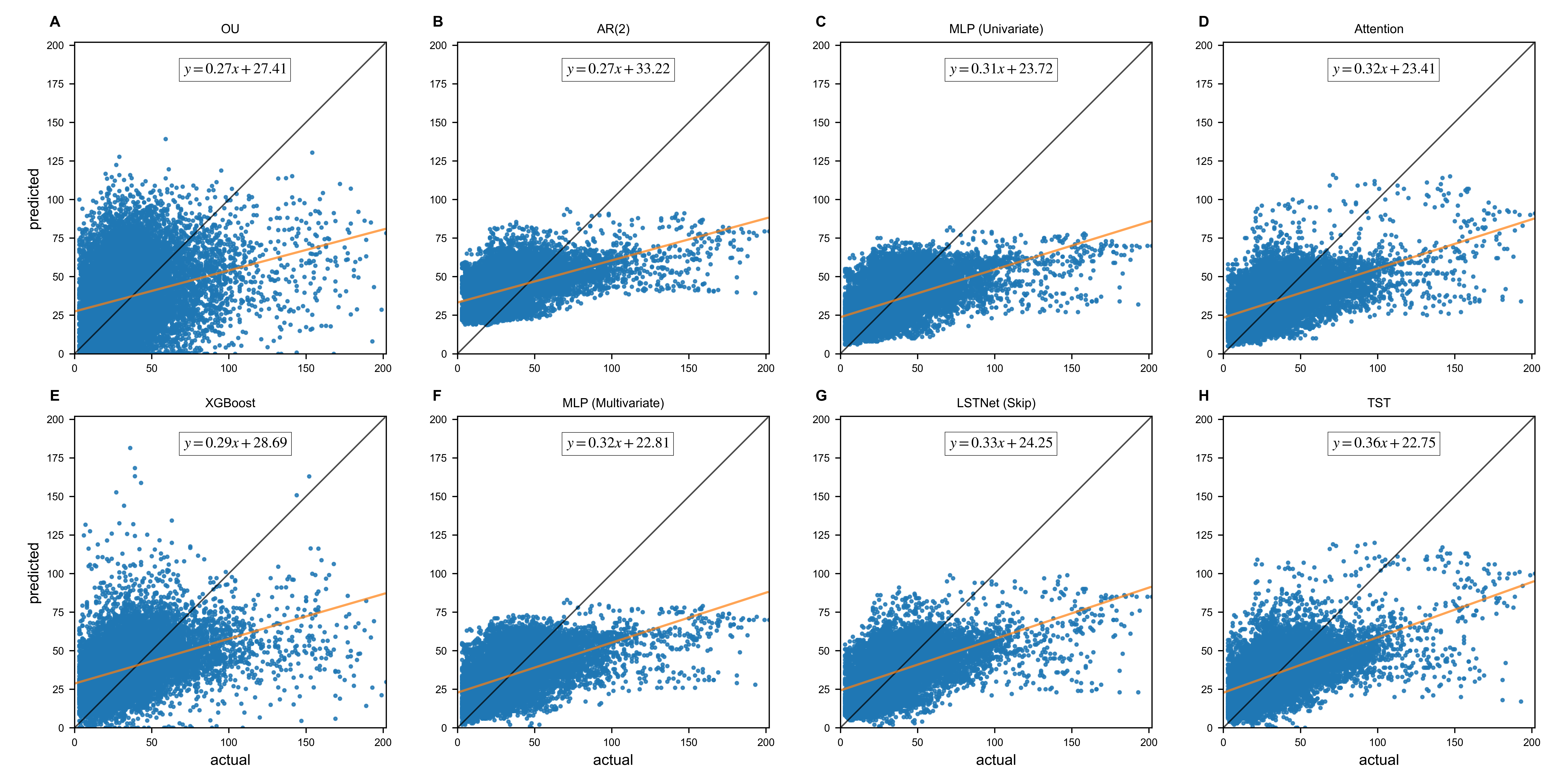}
	\caption{Scatter plot of the result of PM\textsubscript{10} models using MCCR loss. Equations give the best fit line.}
	\label{fig:scatter-mccr-PM10}
\end{figure}

\begin{figure}[!htb]
	\centering
	\includegraphics[width=\textwidth]{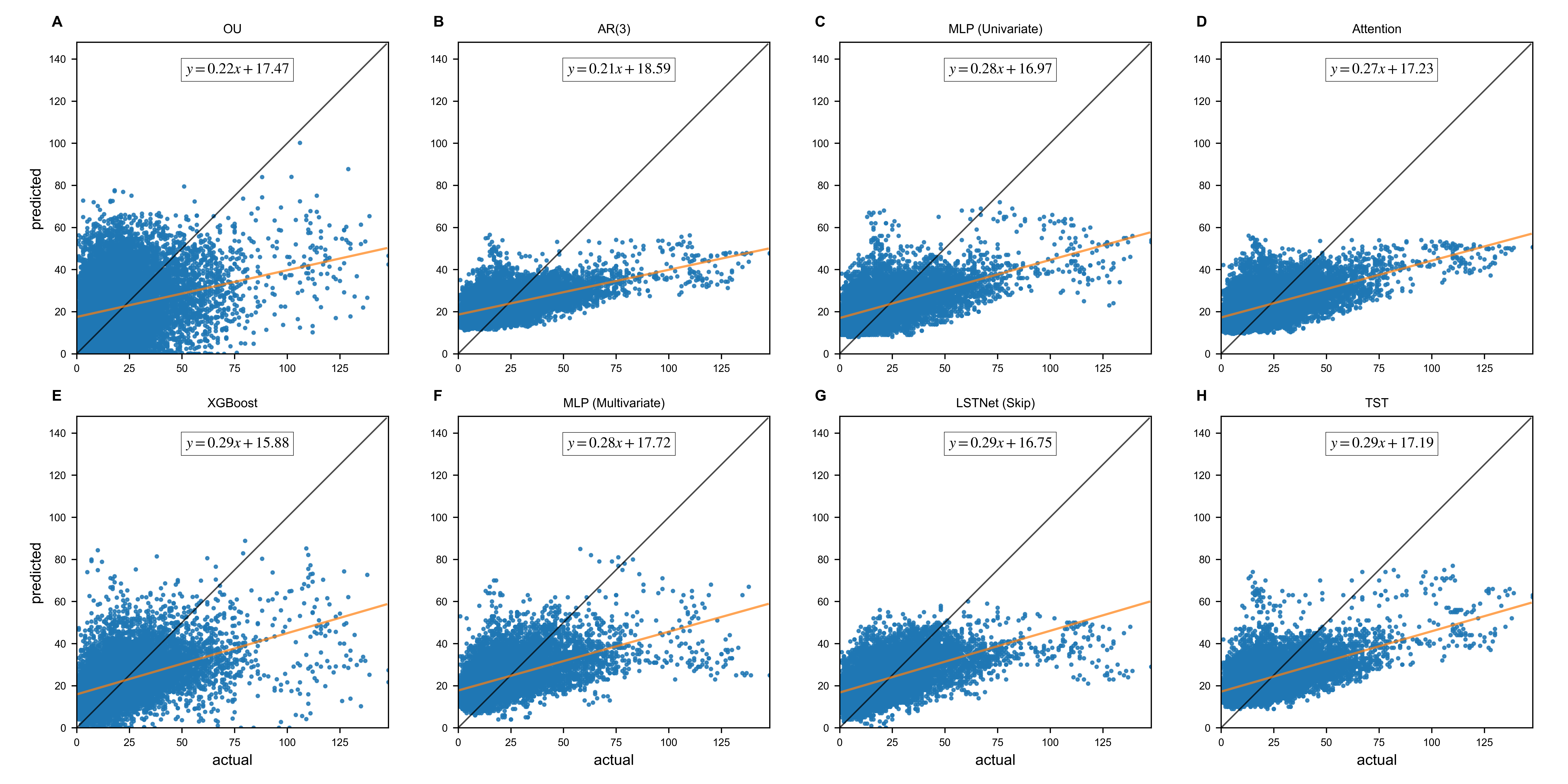}
	\caption{Scatter plot of the result of PM\textsubscript{2.5} models using MSE loss. Equations give the best fit line.}
	\label{fig:scatter-mse-PM25}

	\includegraphics[width=\textwidth]{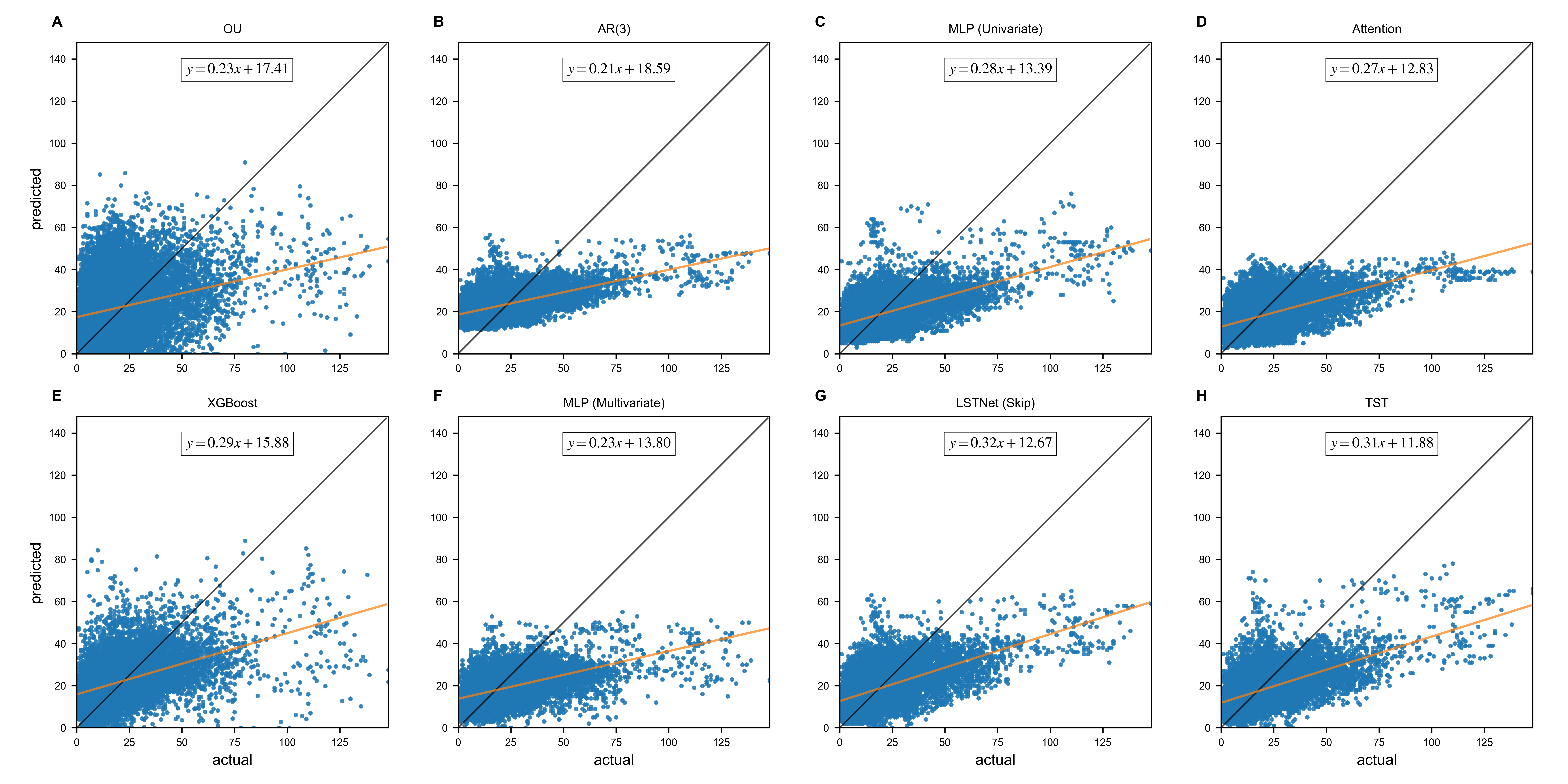}
	\caption{Scatter plot of the result of PM\textsubscript{2.5} models using MCCR loss. Equations give the best fit line.}
	\label{fig:scatter-mccr-PM25}
\end{figure}

\Cref{fig:scatter-mse-PM10,fig:scatter-mse-PM25,fig:scatter-mccr-PM10,fig:scatter-mccr-PM25}
show scatter plots for the 24 h horizon for multiple models and loss functions.
The OU, ARIMA [AR(2) or AR(3)], and XGBoost models are used as baseline models for comparison.
The OU process is a completely stochastic random process.
The ARIMA model uses only a linear equation, and XGBoost is based on multiple decision trees.
The OU process and XGBoost show high dispersion but small bias
for all the loss functions and targets compared to the other models,
indicating incorrect prediction results.
The ARIMA model is highly biased. The OU, ARIMA, and XGBoost models
do not use a gradient-descent-like method or can select only predefined loss functions
such as the MSE loss.

Regardless of the loss function,
the univariate attention model and multivariate models (LSTNet and TST)
show better performance than the MLP models.
The MLP models outperform OU, ARIMA, and XGBoost.
This result is shown qualitatively by the scatter plots
(\Cref{fig:scatter-mse-PM10,fig:scatter-mse-PM25,fig:scatter-mccr-PM10,fig:scatter-mccr-PM25}) and quantitatively
by the metrics (\Cref{tab:result-PM10,tab:result-PM25}).
The results are not surprising, because the LSTNet and transformer models are considered
state-of-the-art models for time series analysis.
From the attention model to the TST model, the number of parameters increases
from thousands to millions, which is larger than the number of training samples
(that is, the models are overparameterized).
By contrast, MLP models have hundreds of parameters and do not overfit.
A recent study showed that overparameterized models do not have large generalization errors
under the assumption of i.i.d. variables \citet{Li2018},
but their use for non-i.i.d. data remains challenging.
However, the empirical result for our data shows that overparameterized models
such as the attention, LSTNet, and TST models are better than conventional or MLP models,
but they exhibit limitations.

Among all the models, the LSTNet and TST models
(bold text in \Cref{tab:result-PM10,tab:result-PM25})
show the best performance overall for quantitative analysis,
as demonstrated by the evaluation metric results presented
in \Cref{tab:result-PM10,tab:result-PM25}.
The univariate attention model shows the best performance
among the univariate models, but for PM\textsubscript{10},
the LSTNet models are better than the attention model,
where the relative difference between them is 5.8\% (12 h horizon)
for the NMAEF and 9.6\% (24 h horizon) for the CORR
when the MSE is used as the loss function.
Using the MCCR loss does not change this tendency.
The MCCR loss of the TST and attention models of
PM\textsubscript{10} is 3.7\% (12 h horizon) for the NMAEF
and 2.4\% for the CORR (12 h horizon). For PM\textsubscript{2.5},
the attention and LSTNet model are the best univariate and multivariate models
according to the MSE loss, which is the same as the PM\textsubscript{10} model result.
Their relative difference is 5.8\% for the NMAEF and 9.7\% for the CORR.

The scatter plots for the 24 h horizon in \Cref{fig:scatter-mse-PM10,fig:scatter-mccr-PM10},
show that the models are highly biased at high PM\textsubscript{10} concentrations,
and only the attention, LSTNet, and TST models outperform the others.
The MSE loss of most models seems to converge to its mean;
thus, the models cannot predict low concentrations.
The multivariate MLP model is unusual
because it is less biased in terms of the MSE loss,
and the predicted values also tend to converge to the mean value,
although the MCCR loss again indicates a highly biased model.
The PM\textsubscript{2.5} models exhibit the same trends as the PM10 models,
and these qualitative trends cannot be measured by the evaluation metrics
in \Cref{fig:scatter-mse-PM25,fig:scatter-mccr-PM25}.
This finding indicates that changing the model improves the model results,
but the improvement is not easily shown by eval-uation metrics.
A comprehensive analysis reveals that state-of-the-art models
such as the LSTNet or TST models outperform the others.
In \citet{Borovykh2019}, a theoretical analysis was performed considering non-i.i.d. variables.
A deep neural network for time series forecasting may be prone to overfitting to noise,
in contrast to networks based on the i.i.d. assumption.
According to these results, to improve the model accuracy,
it is necessary to change other factors, such as the loss function,
not simply the parame-ters or model complexity.

\begin{figure}[!htb]
	\centering

	\includegraphics[width=0.5\textwidth]{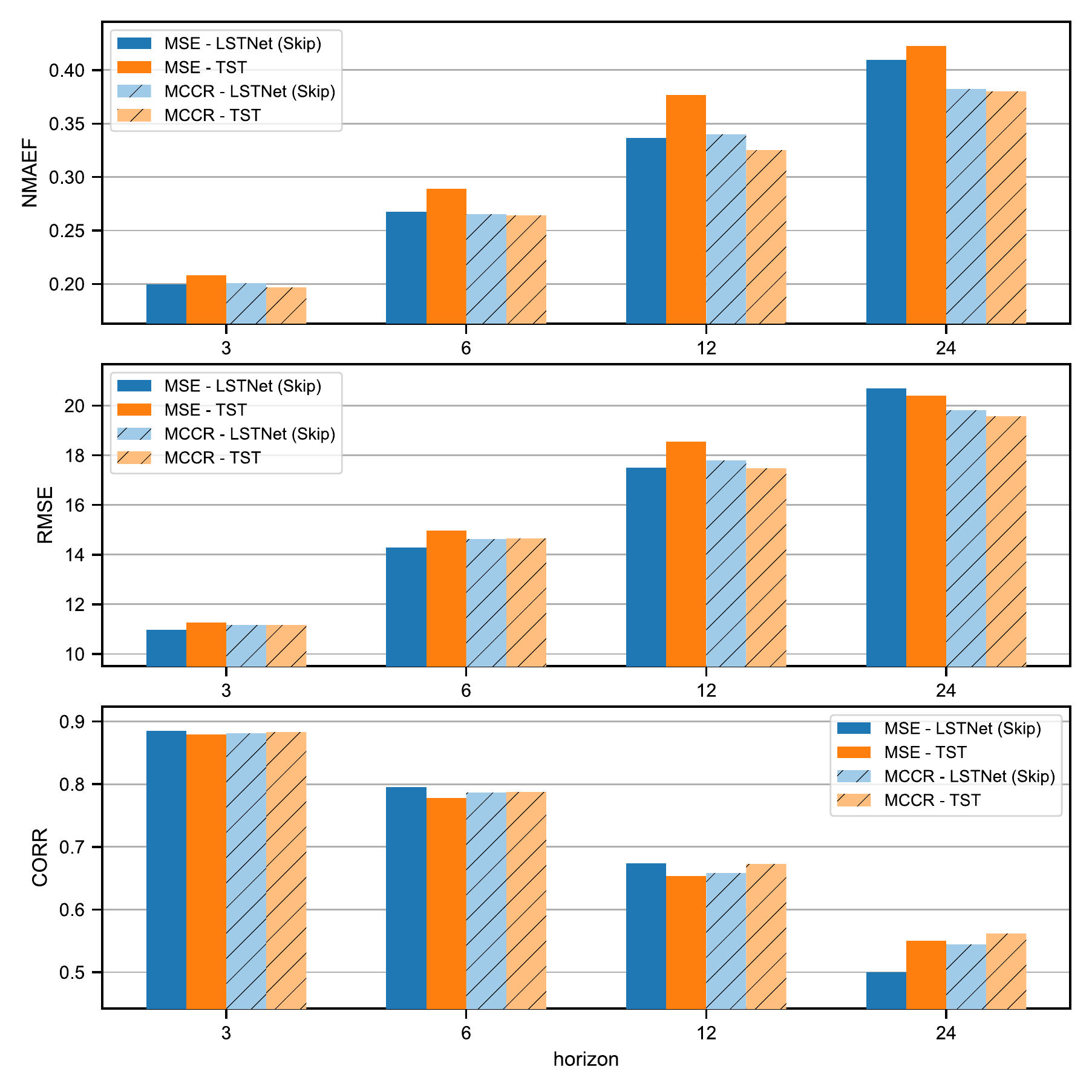}
	\caption{Comparison of RMSE, CORR, and NMAEF for different horizons
		for LSTNet and TST models PM\textsubscript{10}
		and the MSE and MCCR loss functions.}
	\label{fig:bar-PM10}

	\includegraphics[width=0.5\textwidth]{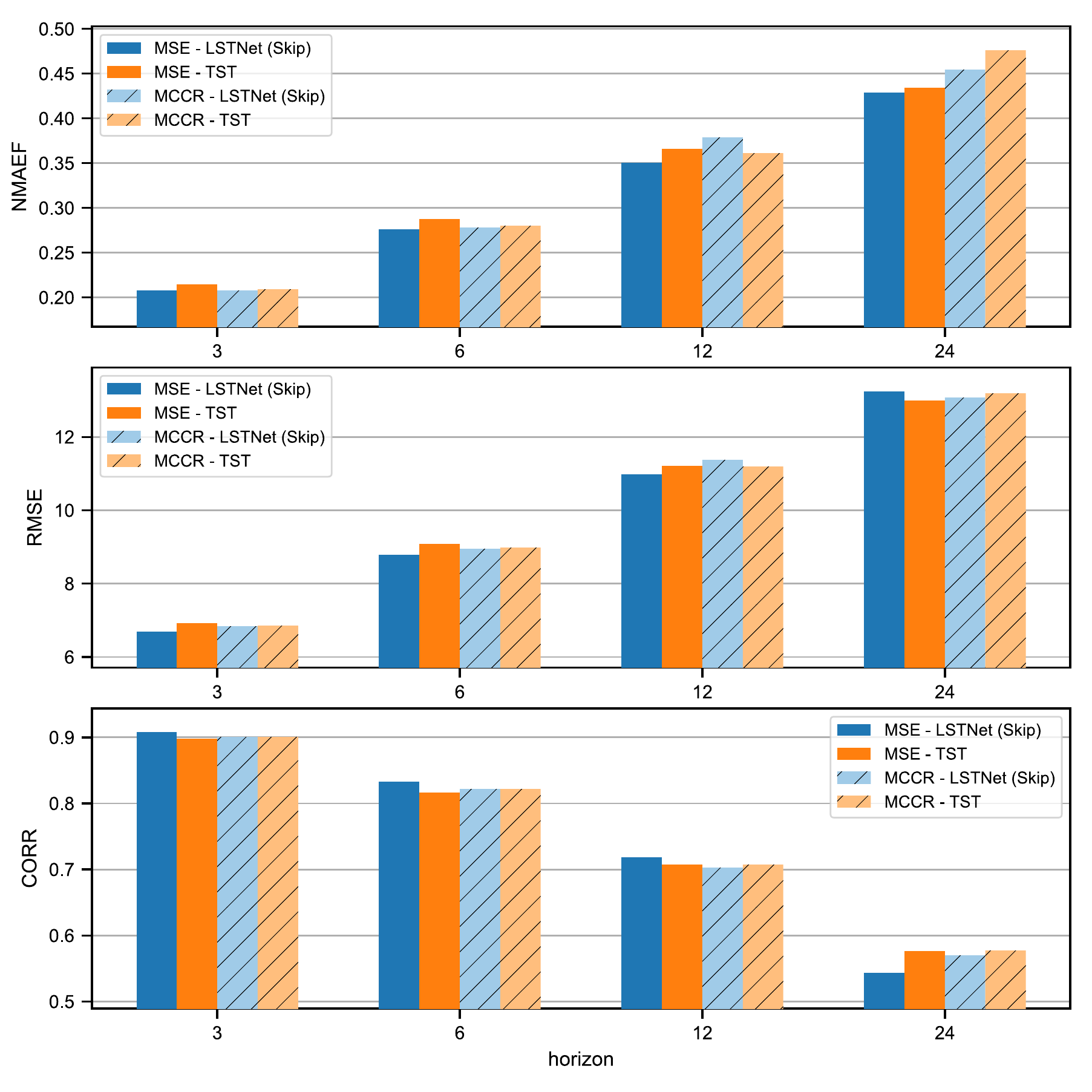}
	\caption{Comparison of RMSE, CORR, and NMAEF for different horizons
		for LSTNet and TST models PM\textsubscript{2.5}
		and the MSE and MCCR loss functions.}
	\label{fig:bar-PM25}
\end{figure}

The MSE loss and MCCR loss are compared in \Cref{fig:bar-PM10,fig:bar-PM25} for various horizons.
Only multivariate models (LSTNet and TST) are used in this context
because they show better results than the others.
The PM\textsubscript{10} models using the MCCR loss outperform those
using the MSE loss for the NMAEF, RMSE, and CORR (5.8\%, 5.8\%, and 8.9\%, respectively).
The PM\textsubscript{2.5} models, however, show better result only
for CORR (8.9\%), whereas the NMAEF is worse (approximately 9.6\%),
and the RMSE is very similar to the NMAE.

\begin{figure}[!htb]
	\centering

	\includegraphics[width=0.7\textwidth]{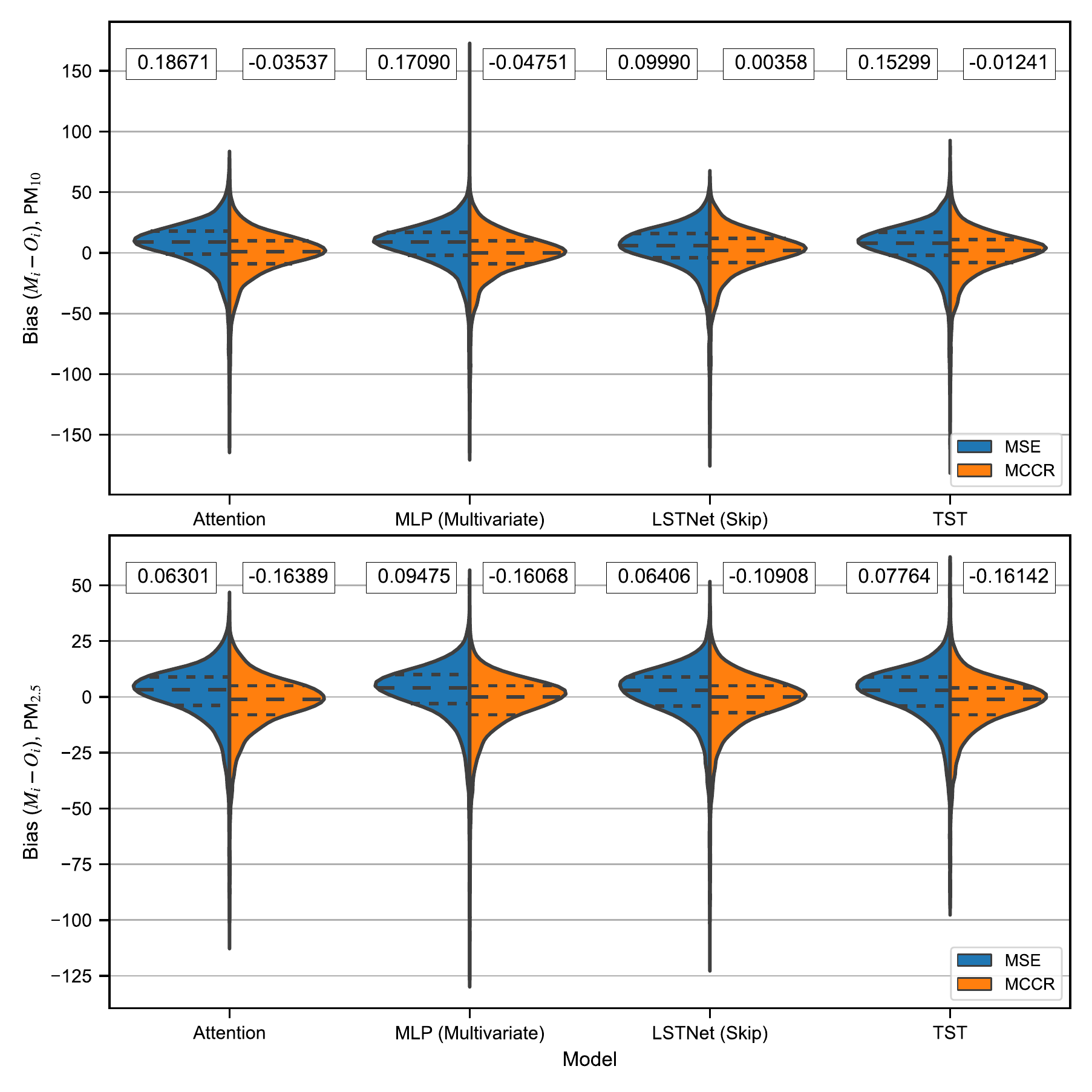}
	\caption{Violin plot of bias ($M_i - O_i$) in 24 h horizon.
		NMBF is displayed at the top of each plot.
		The short-dashed lines indciates first and third quantiels,
		and the long-dashed line indicates the meidan value.}
	\label{fig:violin}
\end{figure}

The bias ($M_i - O_i$) between two the loss functions
at the 24 h horizon is plotted in \autoref{fig:violin}.
The median MSE loss has a positive value, whereas the MCCR loss is close to zero.
This result is consistent with the scatter plots,
as the mean convergence behavior results in a large bias.
In addition, these tendencies are not found in the NMBF values
and the shape of the scatter plot.

\begin{figure}[!htb]
	\centering

	\includegraphics[width=0.7\textwidth]{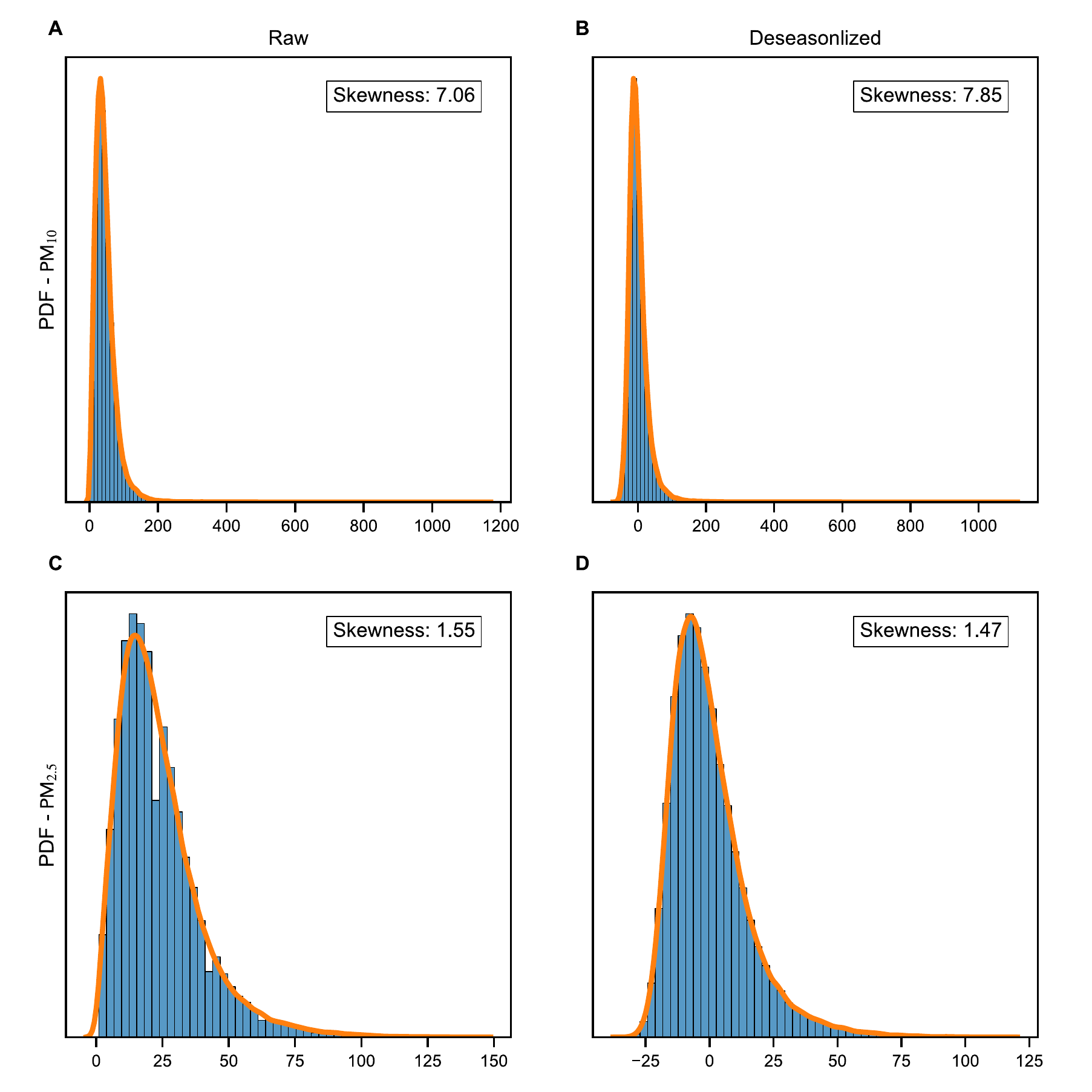}
	\caption{Histogram (blue bars) and KDE plot (orange lines) of raw and
		deseasonalized data distribution.}
	\label{fig:pdf}
\end{figure}

Mean convergence occurs because of the difference
in the distributions of PM\textsubscript{10} and PM\textsubscript{2.5}.
As shown in \autoref{fig:ccdf},
PM\textsubscript{2.5} is not fully fitted by a log-normal distribution.
This fact is also clearly shown in the histogram
and probability density function (PDF) plot obtained
by kernel density estimation (KDE) in \autoref{fig:pdf}.
Unlike that of the PM\textsubscript{10} data,
the distribution of the PM\textsubscript{2.5} data
is similar to a normal distribution.
As mentioned in \citet{Qi2020},
the use of the MCCR loss did not address the problem of KL divergence,
which sacrifices accuracy when choosing the loss function.
When the horizon increases,
the mean convergence became severe (\Cref{fig:horizon-PM10,fig:horizon-PM25}),
even when the MCCR loss was used.
The MCCR loss reduces the severity of this behavior,
but it cannot completely remove it.

Moreover, the prediction results are shifted from the actual values on the time axis.
When the horizon is longer, the shift seems to be larger.
This phenomenon is known as time-warping or time distortion,
and it is often studied
in time series similarity analysis and time series classification \citet{Yi1998}.
To the best of our knowledge, time warping in time series forecasting
is an unexplored topic and should studied further.

\begin{table}
	\centering

	\caption{Summary of results for all models
		PM\textsubscript{10} with 48 h of input MCCR or MSE loss.
		Each row shows the results for a metric and model,
		whereas each column compares the loss function and horizon size.
		Bold text indicates that the value is highest for that metric, horizon size, and loss function.
		The metrics of OU, AR(2) and XGBoost are the same for both loss functions because they are not affected
		by the loss function or cannot be changed.}

	\label{tab:result-PM10}

	\adjustbox{max width=\textwidth}{
	\small
	\begin{tabular}{clrrrrrrrr}
	\toprule
	\multicolumn{2}{c}{Loss function}		& \multicolumn{4}{c}{MSE} 						& \multicolumn{4}{c}{MCCR} 						\\  \midrule
	\multicolumn{2}{c}{Horizon size} 		& \multicolumn{1}{c}{3} & \multicolumn{1}{c}{6} & \multicolumn{1}{c}{12} & \multicolumn{1}{c}{24}  & \multicolumn{1}{c}{3} & \multicolumn{1}{c}{6} & \multicolumn{1}{c}{12} & \multicolumn{1}{c}{24} \\ \midrule
	\multicolumn{1}{c}{Metric} & \multicolumn{1}{c}{Model} & \multicolumn{4}{c}{} 							& \multicolumn{4}{c}{} 			\\ \midrule
	\multirow{8}{*}{NMAEF}	& OU			& 0.31160 	& 0.43167 	& 0.53694 	& 0.61026 	& 0.31160 	& 0.43167 	& 0.53694 	& 0.61026 	\\
							& AR(2)			& 0.20658 	& 0.28436 	& 0.36907 	& 0.44366 	& 0.20658 	& 0.28436 	& 0.36907 	& 0.44366 	\\
							& MLP (Uni.)	& 0.31160 	& 0.43167 	& 0.53694 	& 0.61026 	& 0.31160 	& 0.43167 	& 0.53694 	& 0.61026 	\\
							& Attention		& 0.20696 	& 0.27927 	& 0.35710 	& 0.43374 	& 0.20000 	& 0.27004 	& 0.33744 	& 0.38906 	\\
							& XGBoost		& 0.22264 	& 0.29401 	& 0.37175 	& 0.44077 	& 0.22264 	& 0.29401 	& 0.37175 	& 0.44077 	\\
							& MLP (Multi.)  & 0.23436 	& 0.29161 	& 0.36583 	& 0.42970 	& 0.20455 	& 0.26496 	& 0.32889 	& 0.39465 	\\
							& LSTNet (Skip)	& \bf{0.19949} 	& \bf{0.26763} 	& \bf{0.33630} 	& \bf{0.40943} 	& 0.20058 	& 0.26478 	& 0.33992 	& 0.38197 	\\
							& TST			& 0.20826 	& 0.28890 	& 0.37691 	& 0.42262 	& \bf{0.19674} 	& \bf{0.26380} 	& \bf{0.32507} 	& \bf{0.37987} 	\\ \midrule
	\multirow{8}{*}{RMSE}	& OU			& 14.62660 	& 20.44270 	& 25.43694 	& 28.72156 	& 14.62660 	& 20.44270 	& 25.43694 	& 28.72156 	\\
							& AR(2)			& 11.23659 	& 14.92856 	& 18.42945 	& 21.02014 	& 11.23659 	& 14.92856 	& 18.42945 	& 21.02014 	\\
							& MLP (Uni.)	& 11.39708 	& 14.91436 	& 18.08831 	& 20.47184 	& 11.47124 	& 15.04905 	& 17.91509 	& 19.92637 	\\
							& Attention		& 11.22181 	& 14.79134 	& 18.18526 	& 20.74852 	& \bf{11.14113} 	& 14.72053 	& 17.81423 	& 19.66952 	\\
							& XGBoost		& 12.05999 	& 15.64123 	& 19.20367 	& 22.67406 	& 12.05999 	& 15.64123 	& 19.20367 	& 22.67406 	\\
							& MLP (Multi.)  & 12.23189 	& 15.31694 	& 19.13562 	& 21.02398 	& 11.30984 	& \bf{14.49950} 	& 17.59554 	& 19.91509 	\\
							& LSTNet (Skip)	& \bf{10.96963} 	& \bf{14.29854} 	& \bf{17.49411}	& 20.69443 	& 11.15474 	& 14.60806 	& 17.79352 	& 19.80228 	\\
							& TST			& 11.25933 	& 14.98114 	& 18.54122 	& \bf{20.41276} 	& 11.15219 	& 14.63372 	& \bf{17.45747} 	& \bf{19.56047} 	\\ \midrule
	\multirow{8}{*}{CORR}	& OU			& 0.80755  	& 0.62727  	& 0.42931  	& 0.27401  	& 0.80755  	& 0.62727   & 0.42931   & 0.27401 	\\
							& AR(2)			& 0.88081  	& 0.78134  	& 0.65179  	& 0.53814  	& 0.88081  	& 0.78134  	& 0.65179  	& 0.53814 	\\
							& MLP (Uni.)	& 0.87526  	& 0.77890  	& 0.65876  	& 0.55302  	& 0.87453  	& 0.77841  	& 0.65252  	& 0.53576 	\\
							& Attention		& 0.88089  	& 0.78308  	& 0.65622  	& \bf{0.55321}  	& 0.88112  	& 0.78211  	& 0.65675  	& 0.55157 	\\
							& XGBoost		& 0.85994  	& 0.75357  	& 0.60975  	& 0.41355  	& 0.85994  	& 0.75357  	& 0.60975  	& 0.41355 	\\
							& MLP (Multi.)  & 0.85841  	& 0.76903  	& 0.62315	& 0.52384	& 0.87692  	& \bf{0.79021}	& 0.66923	& 0.54006 	\\
							& LSTNet (Skip)	& \bf{0.88502}  	& \bf{0.79557}  	& \bf{0.67414}  	& 0.49977	& 0.88086  	& 0.78654 	& 0.65783 	& 0.54404 	\\
							& TST			& 0.87931  	& 0.77791  	& 0.65392	& 0.54980	& \bf{0.88282}  	& 0.78749	& \bf{0.67248}	& \bf{0.56119} 	\\ \midrule
	\multirow{8}{*}{NMBF}	& OU			& \bf{-0.00323} 	& \bf{-0.00309} 	& \bf{-0.00028} 	& \bf{0.00463} 	& \bf{-0.00323} 	& \bf{-0.00309} 	& \bf{-0.00028} 	& 0.00463	\\
							& AR(2) 		& 0.04033	& 0.07405 	& 0.12661 	& 0.19097 	& 0.04033 	& 0.07405 	& 0.12661 	& 0.19097	\\
							& MLP (Uni.)	& 0.01702	& 0.06066	& 0.10342	& 0.16357	& -0.00163	& 0.00440	& -0.03813	& -0.03663	\\
							& Attention		& 0.03545	& 0.06052	& 0.10803	& 0.18671	& -0.00687	& -0.01279	& -0.02345	& -0.03537	\\
							& XGBoost		& 0.02536	& 0.04296	& 0.07382	& 0.08280	& 0.02536	& 0.04296	& 0.07382	& 0.08280	\\
							& MLP (Multi.)	& 0.05466	& 0.07977	& 0.12821	& 0.17090	& -0.00287	& -0.01381	& -0.01395	& -0.04751	\\
							& LSTNet (Skip)	& 0.01764	& 0.03158	& 0.05536	& 0.09990	& 0.01848	& 0.02684	& -0.02924	& \bf{0.00358}	\\
							& TST			& 0.02888	& 0.06234	& 0.14025	& 0.15299	& 0.01167	& 0.01944	& -0.00590	& -0.01241 	\\
	\bottomrule
	\end{tabular}
	}
\end{table}

\begin{table}
	\centering

	\caption{Summary of results for all models
		PM\textsubscript{2.5} with 48 h of input MCCR or MSE loss.
		Each row shows the results for a metric and model,
		whereas each column compares the loss function and horizon size.
		Bold text indicates that the value is highest for that metric, horizon size, and loss function.
		The metrics of OU, AR(3) and XGBoost are the same for both loss functions because they are not affected
		by the loss function or cannot be changed.}

	\label{tab:result-PM25}

	\adjustbox{max width=\textwidth}{
	\small
	\begin{tabular}{clrrrrrrrr}
	\toprule
	\multicolumn{2}{c}{Loss function}		& \multicolumn{4}{c}{MSE} 						& \multicolumn{4}{c}{MCCR} 						\\  \midrule
	\multicolumn{2}{c}{Horizon size} 		& \multicolumn{1}{c}{3} & \multicolumn{1}{c}{6} & \multicolumn{1}{c}{12} & \multicolumn{1}{c}{24}  & \multicolumn{1}{c}{3} & \multicolumn{1}{c}{6} & \multicolumn{1}{c}{12} & \multicolumn{1}{c}{24} \\ \midrule
	\multicolumn{1}{c}{Metric} & \multicolumn{1}{c}{Model} & \multicolumn{4}{c}{} 							& \multicolumn{4}{c}{} 			\\ \midrule
	\multirow{8}{*}{NMAEF}	& OU 			& 0.33740 	& 0.46857 	& 0.60018 	& 0.68653 	& 0.33967 	& 0.46773 	& 0.59131 	& 0.68730 	\\
							& AR(3) 		& 0.21508 	& 0.29414 	& 0.37558 	& 0.44522 	& 0.21508 	& 0.29414 	& 0.37558 	& 0.44522 	\\
							& MLP (Uni.) 	& 0.21828 	& 0.29117 	& 0.36862 	& 0.43095 	& 0.21753 	& 0.29193 	& 0.37206 	& 0.46097 	\\
							& Attention 	& 0.22568 	& 0.29492 	& 0.36727 	& 0.43198 	& 0.21453 	& 0.29735 	& 0.39183 	& 0.48990 	\\
							& XGBoost 		& 0.23096 	& 0.29453 	& 0.37795 	& 0.45343 	& 0.23096 	& 0.29453 	& 0.37795 	& \bf{0.45343}	\\
							& MLP (Multi.) 	& 0.22659 	& 0.29086 	& 0.36495 	& 0.43857 	& 0.20924 	& 0.27786 	& 0.39393 	& 0.48344	\\
							& LSTNet (Skip) & \bf{0.20757} 	& \bf{0.27602} 	& \bf{0.35062} 	& \bf{0.42858} 	& \bf{0.20747} 	& \bf{0.27770} 	& 0.37823 	& 0.45436	\\
							& TST 			& 0.21458 	& 0.28792 	& 0.36576 	& 0.43434 	& 0.20901 	& 0.28004 	& \bf{0.36093} 	& 0.47585	\\ \midrule
	\multirow{8}{*}{RMSE}	& OU			& 9.42310	& 13.10905	& 16.74847	& 19.28485	& 9.47383	& 13.16762	& 16.65591	& 19.33624	\\
							& AR(3)			& 6.93497	& 9.20198	& 11.48500	& 13.39068	& 6.93497	& 9.20198	& 11.48500	& 13.39068	\\
							& MLP (Uni.)	& 7.05789	& 9.25539	& 11.44019	& 13.16623	& 6.97408	& 9.18980	& 11.30530	& 13.16732	\\
							& Attention		& 7.33964	& 9.37080	& 11.41769	& 13.06500	& 6.92724	& 9.22089	& 11.56166	& 13.56691	\\
							& XGBoost		& 7.55752	& 9.37789	& 11.99462	& 14.24905	& 7.55752	& 9.37789	& 11.99462	& 14.24905	\\
							& MLP (Multi.)	& 6.99418	& 9.05431	& 11.32332	& 13.45415	& \bf{6.70583}	& \bf{8.78049}	& 11.81080	& 13.97028	\\
							& LSTNet (Skip)	& \bf{6.69026}	& \bf{8.78158}	& \bf{10.98092}	& 13.24178	& 6.83660	& 8.94400	& 11.36367	& \bf{13.07116}	\\
							& TST			& 6.92550	& 9.08245	& 11.20545	& \bf{12.99081}	& 6.84600	& 8.98041	& \bf{11.18389}	& 13.17671	\\ \midrule
	\multirow{8}{*}{CORR}	& OU			& 0.81975	& 0.65019	& 0.42922	& 0.23856	& 0.81864	& 0.65106	& 0.43967	& 0.24084	\\
							& AR(3)			& 0.89910	& 0.81549	& 0.69832	& 0.56058	& 0.89910	& 0.81549	& 0.69832	& 0.56058	\\
							& MLP (Uni.)	& 0.89364	& 0.80890	& 0.68903	& 0.55225	& 0.89687	& 0.81209	& 0.69759	& 0.56876	\\
							& Attention		& 0.88970	& 0.80709	& 0.69212	& 0.56860	& 0.89857	& 0.81279	& 0.68962	& 0.54083	\\
							& XGBoost		& 0.87710	& 0.80211	& 0.64844	& 0.45717	& 0.87710	& 0.80211	& 0.64844	& 0.45717	\\
							& MLP (Multi.)	& 0.90058	& 0.82361	& 0.70043	& 0.53116	& 0.90481	& 0.82975	& 0.68050	& 0.49566	\\
							& LSTNet (Skip)	& \bf{0.90760}	& \bf{0.83255}	& \bf{0.71840}	& 0.54388	& \bf{0.90010}	& \bf{0.82169}	& 0.70240	& 0.56993	\\
							& TST			& 0.89831	& 0.81675	& 0.70713	& \bf{0.57669}	& 0.90009	& 0.82127	& \bf{0.70663}	& \bf{0.57703}	\\ \midrule
	\multirow{8}{*}{NMBF}	& OU			& \bf{-0.00202}	& -0.00720	& \bf{-0.00745}	& \bf{0.00006}	& \bf{0.00275}	& \bf{0.00527}	& \bf{0.00916}	& \bf{0.00181}	\\
							& AR(3)			& 0.01556	& 0.02749	& 0.04589	& 0.06797	& 0.01556	& 0.02749	& 0.04589	& 0.06797	\\
							& MLP (Uni.)	& 0.02095	& 0.02734	& 0.05274	& 0.05668	& -0.02503	& -0.03331	& -0.04715	& -0.11779	\\
							& Attention		& 0.01346	& 0.02096	& 0.03643	& 0.06301	& -0.02102	& -0.04517	& -0.09404	& -0.16389	\\
							& XGBoost		& -0.00995	& \bf{0.00255}	& 0.01059	& 0.02108	& -0.00995	& 0.00255	& 0.01059	& 0.02108	\\
							& MLP (Multi.)	& 0.06821	& 0.07098	& 0.07403	& 0.09475	& -0.00009	& -0.02031	& -0.10505	& -0.16068	\\
							& LSTNet (Skip)	& 0.01879	& 0.01997	& 0.03186	& 0.06406	& -0.00422	& -0.00659	& -0.08228	& -0.10908	\\
							& TST			& 0.01678	& 0.01718	& 0.04882	& 0.07764	& -0.01225	& -0.01702	& -0.02728	& -0.1614	\\
	\bottomrule
	\end{tabular}
	}
\end{table}

\section{Conclusion}
\label{sec:conclusion}

We presented a novel TST-model-based air pollution fore-casting model.
We also showed that the prediction result can be improved
by simply varying the loss function depending on the data distribution.
The proposed approach exhibits not only better performance but also simplicity.
Air pollution data such as PM\textsubscript{10} and PM\textsubscript{2.5} data
involve multiple factors and highly complex seasonality.
The consideration of complex seasonality for different time scales,
such as daily or hourly, was also proposed.
Our results suggest that setting the input length to 48 h,
using multivariate models, applying state-of-the-art models such as LSTNet or TST,
and using the MCCR loss function provides the best results.
Importantly, our results provide further evidence
for the importance of the data distribution in machine learning research.
If the data characteristics are not considered,
even state-of-the-art models cannot give better results.

Future research should focus on the development
of methods of handling long-range dependence and extreme values.
As mentioned in Section 3.4, multiple approaches have been proposed recently
to train memory over a longer horizon on the basis
of the well-established extreme value theory \citet{Ding2019,Ribeiro2020}.

In addition, theoretical research on the generalization of non-i.i.d. data
such as time series data is also needed.
Time series forecasting is known as a challenging area
owing to the possibility of overfitting the dataset.
In \citet{Borovykh2019}, the generalization capabilities of fully connected neural networks
were extensively analyzed using a Hessian matrix.
A thorough search of the relevant literature showed
that this is the only paper to provide a theoretical foundation
for the generalization of non-i.i.d. data.
Time series data are a classic example of non-i.i.d. values,
and noise is easily overfitted to training data with an overparameterized model.
Moreover, a difference in distributions between the training and test data
might degrade the out-of-sample prediction performance.
Although insights on the generalization of time series models are presented in \citet{Borovykh2019},
studies of more complex models such as RNNs are needed.
In addition, as mentioned in \autoref{subsec:performance},
the generality of time warping in time series should be investigated further.

\section*{Acknowledgments}
This work was supported by a National Research Foundation of Korea (NRF)
grant funded by the Korean government (MSIP) (2017R1E1A1A03070282).


\nomenclature{$x(t)$}{Single input time series}
\nomenclature{$s_{y,\textrm{smoothed}}$}{Smoothed yearly seasonality}
\nomenclature{$s_{w}$}{Weekly seasonality}
\nomenclature{$s_{h}$}{Daily seasonality}
\nomenclature{$\textrm{res}_{h}$}{Residuals}
\nomenclature{$C(r)$}{Autocorrelation}
\nomenclature{$V(s)$}{DFA fluctuation function}
\nomenclature{$h$}{DFA fluctuation exponent}
\nomenclature{$\xi$}{Long-range dependence power law exponent}
\nomenclature{$\alpha$}{Pareto Index}
\nomenclature{$\bar{F}$}{Complementary Cumulative Distribution Function}
\nomenclature{$F$}{Cumulative Distribution Function}
\nomenclature{$\beta$}{MCCR scale parameter}
\nomenclature{$Y_t$}{Actual value}
\nomenclature{$\hat{Y}_t$}{Predicted value}
\nomenclature{$O_i$}{Actual value}
\nomenclature{$M_i$}{Predicted value}

\printnomenclature

\bibliographystyle{unsrtnat}
\bibliography{references}  

\end{document}